%% file: main.tex
\documentclass[nonacm,screen,sigplan]{acmart}

\AtBeginDocument{%
  \providecommand\BibTeX{{%
    \normalfont B\kern-0.5em{\scshape i\kern-0.25em b}\kern-0.8em\TeX}}}

\setcopyright{acmcopyright}
\copyrightyear{2018}
\acmYear{2018}
\acmDOI{10.1145/1122445.1122456}

\acmJournal{POMACS}
\acmVolume{37}
\acmNumber{4}
\acmArticle{111}
\acmMonth{8}
\usepackage{graphicx}
\graphicspath{plots/}
\usepackage{tabularx}
\usepackage{longtable}
\usepackage{forest}
\usepackage{hyperref}

\usepackage{tikz}

\begin{document}

\title{Neural Natural Language Processing for Unstructured Data in Electronic Health Records: a Review}

\author{Irene Li, Jessica Pan, Jeremy Goldwasser, Neha Verma, Wai Pan Wong}
\author{Muhammed Yavuz Nuzumlalı, Benjamin Rosand, Yixin Li, Matthew Zhang, David Chang}
\author{R. Andrew Taylor, Harlan M. Krumholz and  Dragomir Radev}

\email{{irene.li,jessica.pan,jeremy. goldwasser,neha.verma,waipan.wong}@yale.edu}

\email{{yavuz.nuzumlali,benjamin.rosand,yixin.li,matthew.zhang,david.chang}@yale.edu
}

\email{{richard.taylor,harlan.krumholz,dragomir.radev}@yale.edu . } 

\affiliation{Yale University, New Haven CT 06511}




\renewcommand{\shortauthors}{Li et al.}

\begin{abstract}

  Electronic health records (EHRs), digital collections of patient healthcare events and observations, are ubiquitous in medicine and critical to healthcare delivery, operations, and research. Despite this central role, EHRs are notoriously difficult to process automatically. Well over half of the information stored within EHRs is in the form of unstructured text (e.g. provider notes, operation reports) and remains largely untapped for secondary use. Recently, however, newer neural network and deep learning approaches to Natural Language Processing (NLP) have made considerable advances, outperforming traditional statistical and rule-based systems on a variety of tasks. In this survey paper, we summarize current neural NLP methods for EHR applications. We focus on a broad scope of tasks, namely,  classification and prediction, word embeddings, extraction, generation, and other topics such as question answering, phenotyping, knowledge graphs, medical dialogue, multilinguality, interpretability, etc.

\end{abstract}

\ccsdesc[500]{General and reference~Surveys and overviews}
\ccsdesc[500]{Computing methodologies~Natural language processing}
\ccsdesc[500]{Computing methodologies~Machine learning algorithms}

\keywords{natural language processing, neural networks, EHR, unstructured data}

\maketitle

\input{0_intro}

\input{1_preliminary}
\input{2_classification}
\input{3_clinical_embeddings}

\input{4_extraction}

\input{5_generation_summarization}
\input{6_other_topics}
\input{8_conclusion}

\bibliographystyle{ACM-Reference-Format}
\bibliography{sample}
\clearpage

\input{appendix}

\end{document}

%% file: 0_intro.tex
\section{Introduction}




Electronic health records (EHRs), digital collections of patient healthcare events and observations, are now ubiquitous in medicine and critical to healthcare delivery, operations, and research  \cite{gunter2005emergence,fjames2015registry,denaxas2015big,cowie2017electronic}.
Data within EHRs are often classified based on collection and representation formats belonging into two classes: structured or unstructured \cite{consultantunstructured}.  Structured EHR data consist of heterogeneous sources like diagnoses, medications, and laboratory values in fixed numerical or categorical fields. Unstructured data, in contrast, refer to free-form text written by healthcare providers, such as clinical notes and discharge summaries. Unstructured data represent about 80\% of total EHR data, but unfortunately remain very difficult to process for secondary use \cite{xiao2018opportunities}. In this survey paper, we focus our discussion mainly on unstructured text data in EHRs and newer neural, deep-learning based methods used to leverage this type of data.

In recent years, artificial neural networks have dramatically impacted fields such as speech recognition, computer vision (CV), and natural language processing (NLP) within medicine and elsewhere \cite{murphy2012machine,chen2020probabilistic}. These networks, which most often are composed of multiple layers in a modeling strategy known as deep learning, have come to outperform traditional rule-based and statistical methods on many tasks. Recognizing the inherent performance advantages and potential to improve health outcomes by unlocking unstructured EHR text data, neural NLP methods  are attracting great interest among researchers in artificial intelligence for healthcare. 

\subsection{Challenges and difficulties}
In this survey, we focus on neural approaches to analyzing clinical texts in EHRs. While these data can carry abundant useful information in healthcare, there also exist significant challenges and difficulties including: 
\begin{itemize}

\item \textit{Privacy.}
Due to the extremely sensitive nature of the information contained in EHRs and the existence of regulatory laws such as the (US) Health Insurance and Portability and Accountability Act (HIPAA), maintenance of privacy within analytic pipelines is imperative. \cite{kushida2012strategies,fernandez2013security}. Therefore, often before any downstream tasks can be performed or any data can be shared with others, additional privacy-preserving steps must be taken. Removing identifying information from a large corpus of EHRs is an expensive process. It is difficult to automate and requires annotators with domain expertise.

\item \textit{Lack of annotations.} 
Many existing machine learning and deep learning models are supervised and thus require labeled data for training. Annotating EHR data can be challenging because of the cognitive complexity of the task and because of the variability in the quality of data. Neural networks require large amounts of text on which to train, and unfortunately, useful EHR data are often in short supply. For some tasks, only qualified annotators can be recruited to complete annotations. Even when annotators are available, the quality of the annotations is sometimes hard to ensure. There may be disagreement among the annotators which can make the evaluations more difficult and controversial.

\item \textit{Interpretability.}
Deep neural networks are able to achieve superior results compared with other methods. However, in many fields, they are often treated as black boxes \cite{zhang2018visual,vellido2019importance}. Typically, a neural network model has a large number of trainable parameters, which causes severe difficulties for model interpretability. Moreover, unlike linear models, which are usually more straightforward and explainable, neural networks consist of non-linear layers and complex architectures further hampering interpretation. Recently, there have been a few attempts to produce explainable deep neural networks to increase model transparency \cite{che2015distilling, che2016interpretable,mullenbach2018medical}.

\end{itemize}

\begin{figure}[t!]
    \centering
    \small
    \scalebox{0.6}{
    \begin{forest}
          for tree={
            line width=0.5pt,
            draw=black,
            fit=rectangle,
            edge={color=black,>={[]}, ->},
            if level=0{%
              l sep+=1cm,
              for descendants={%
                calign=first,
              },
              align=center,
              parent anchor=south,
            }{%
              if level=1{%
                parent anchor=south west,
                child anchor=north,
                tier=three ways,
                align=center,
                for descendants={%
                  child anchor=west,
                  parent anchor=west,
                  align=left,
                  anchor=west,
                  xshift=-20pt,
                  edge path={
                    \noexpand\path[\forestoption{edge}]
                    (!to tier=three ways.parent anchor) |-
                    (.child anchor)\forestoption{edge label};
                  },
                },
              }{}%
            },
          }
          [Neural NLP Methods for EHR Tasks
            [Classification
              [Medical\\Text\\Classification
                [Segmentation
                    [Word Sense\\ Disambiguation
                        [Medical Coding
                            [Medical Outcome\\Prediction
                                [De-identification]
                            ]
                        ]
                    ]
                ]
              ]
            ]
            [Embeddings
              [Medical\\Concept\\Embeddings
                [Visiting\\Embeddings
                    [Patient\\Embeddings
                        [BERT-based\\Embeddings]
                    ]
                ]
              ]
            ]
            [Extraction
              [Named Entity\\Recognition
                [Entity Linking
                    [Relation and \\
                    Event Extraction
                        [Medication\\Information\\Extraction]
                    ]
                ]
              ]
            ]
            [Generation
              [EHR Generation
                [Summarization
                    [Medical\\Language\\Translation]
                ]
              ]
            ]
            [Other Topics
              [Question Answering
                [Phenotyping
                    [Knowledge Graphs
                        [Medical Dialogues
                            [Multilinguality
                                [Interpretability
                                    [Applications in\\Public Health]
                                ]
                            ]
                        ]
                    ]
                ]
              ]
            ]
          ]
        \end{forest}
        }
\caption{Various tasks covered in this survey.}
\label{fig:taxonomy}
\end{figure}
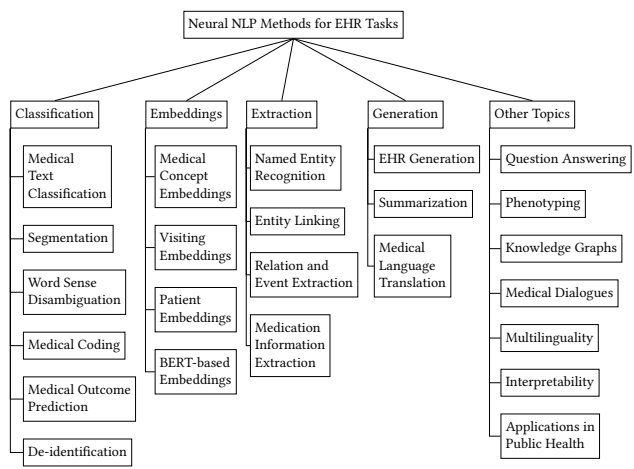
\subsection{Related Work} 

Prior surveys have focused on a variety of deep learning topics within health informatics, bioinformatics, including EHRs.
An early survey by Miotto et al. \cite{Riccardo2017deep} summarized deep learning methods for healthcare and their applications in clinical imaging, EHRs, genomics, and mobile domains. The authors reviewed a number of tasks including predicting diseases, modeling phenotypes, and learning representations of medical concepts, such as diseases and medications. Some other survey papers, such as \cite{Shickel2018deep,al-Aiad2018survey} described clinical applications that use deep learning techniques including information extraction, representation learning, outcome prediction, phenotyping and de-identification. Similarly, a systematic survey by Xiao et al. \cite{xiao2018opportunities} targeted five categories: disease detection/classification, sequential prediction, concept embedding, data augmentation and EHR data privacy. 
Kwak and Hui \cite{kwak2019deephealth} reviewed research applying artificial intelligence to health informatics. Specifically in the EHR field, they included the following applications: outcome prediction,  computational phenotyping,  knowledge extraction, representation learning, de-identification and medical intervention recommendations. Assale et al. \cite{assale2019revival} presented a literature review of how free-text content from electronic patient records can benefit from recent Natural Language Processing techniques, selecting four application domains: data quality, information extraction, sentiment analysis and predictive models, and automated patient cohort selection. 
Another recent survey \cite{wu2019deepmethod} summarized deep learning methods in clinical NLP. The authors reviewed methods such as deep learning architectures, embeddings and medical knowledge on four groups of clinical NLP tasks: text classification, named entity recognition, relation extraction, and others. Joshi et al. \cite{epidemic_survey} discussed textual epidemic intelligence or the detection of disease outbreaks via medical and informal (i.e. the Web) text. 

\subsection{Motivation} 
While the aforementioned reviews target different applications or techniques, and vary on the scope of selected topics, in this survey, we look at a more comprehensive coverage of applications and tasks, and we include several works with very recent BERT-based models. Specifically, we summarize a broad range of existing literature on deep learning methods with clinical text data. Most of the papers on clinical NLP discussed achieve performance at or near the state-of-the-art for their tasks. 

We now provide some NLP preliminaries, and then we explore deep learning methods on a various tasks: classification, embeddings, extraction, generation, and other topics including question answering, phenotyping, knowledge graphs, and more. Figure ~\ref{fig:taxonomy} shows the scope of the tasks. Finally, we also summarize relevant datasets and existing tools.

%% file: 1_preliminary.tex
\section{Preliminaries}
In this section, we present an overview of important concepts in deep learning and natural language processing. In particular, we cover commonly used deep learning architectures in NLP, word embeddings, recent Transformer models, and concepts from transfer learning.

\subsection{Basic Architectures}
We first summarize some basic network architectures commonly used in NLP. We introduce autoencoders, convolutional neural networks, recurrent neural networks, and sequence-to-sequence models.

\subsubsection{Autoencoders}



Autoencoders \cite{kramer1991nonlinear} learn lower dimensional representations of a given input, while preserving salient features. The autoencoder is a neural network that encodes its input into (typically) fewer dimensions, and then decodes this representation to reconstruct the original input. In other words, the autoencoder identifies the hidden semantic features that can be used to represent the input data. 


Several variations of the autoencoder exist. The denoising autoencoder \cite{vincent2008extracting} corrupts each input by randomly setting some of the input values to zero, but still trains to reconstruct the uncorrupted sample. Doing so prevents overfitting and forces the model to more robustly learn the dependencies between input features. 

Another modification, the variational autoencoder (VAE) \cite{kingma2014autoencoding}, learns to represent each sample with a probability distribution rather than a fixed encoding. This increases model robustness by accounting for greater variance in the latent space. And, because a VAE learns a latent state distribution, its decoder can also generate new samples by randomly sampling encodings from the representation space. Besides, such a latent state distribution forces the underlying manifold or topology to be continuous, making it suitable and important for density-based clustering and topic modeling. 

A third variation is the stacked autoencoder \cite{vincent2010stacked}. Rather than a single autoencoder, this model consists of a chain of successive units. Each autoencoder layer within a stacked autoencoder inputs and reconstructs the encoding layer from the previous autoencoder. Stacked denoising autoencoders are used in a number of NLP applications such as sentiment analysis \cite{zhang2015japanese}.

\subsubsection{Convolutional Neural Networks}


Convolutional Neural Networks (CNNs) \cite{fukushima1982neocognitron,waibel1989phoneme,lecun1998gradient} are a classic neural model based on the human visual cortex. Each layer of a CNN convolves input matrices with smaller filters whose parameters are learned by the network. In effect, it will learn higher  and higher level features with each successive layer. In image processing, it may detect edges in the first layers, piece together those low-level shapes in subsequent layers, and so on until it can identify abstract concepts like faces or specific objects. 
CNNs can also be applied to common NLP tasks, such as text classification \cite{kimconvolutional,zhang2015character}.


\subsubsection{Recurrent Neural Networks}

Recurrent Neural Networks (RNNs) \cite{werbos1988generalization,robinson1987static} process sequential input at discrete time steps. Because they share the same weights at every step, they are able to handle variable-length inputs; this property makes RNNs very helpful in NLP applications, which have sequential text input \cite{werbos1990backpropagation,hopfield1982neural,mozer1989a}. In the standard (or \lq\lq Vanilla\rq\rq) RNN, each time step has a cell that processes both the input at that step and the processed input from all preceding steps. From these inputs, each cell generates an output that gets passed to the next RNN cell, and another (optional) output that can be used to make a decision at that time step.


Vanilla RNNs struggle to learn long-term temporal dependencies because their gradients can grow or decay exponentially over multiple time steps. To solve this issue, the Vanilla RNN cell can be replaced by a Long Short-Term Memory (LSTM) cell \cite{hochreiter1997long}, a Bidirectional Long Short-Term Memory (BiLSTM) \cite{graves2005bid} cell or a Gated Recurrent Unit (GRU) cell \cite{cho2014learning}. These cells have gates that control the flow of information, making the network better at retaining memory over multiple time steps. Such RNN-based models have been widely applied in tasks including text classification \cite{liu2016recurrent} and language understanding \cite{yao2013recurrent}.

\subsubsection{Sequence-to-sequence models}

Each cell in an RNN can produce an output in addition to a hidden state. This gives them a large amount of flexibility: they can convert one input to one output, one input to many outputs, many inputs to one output, or many inputs to many outputs. Sequence-to-sequence (seq2seq) models are a special kind of many-to-many RNN, in which the whole input sequence is encoded before an output sequence is decoded one token at a time \cite{sutskever2014sequence}. This presents an encoder-decoder framework resembling that of the autoencoder, but without the objective for the decoder to output the original input. At each decoder time step, the seq2seq model generates a token in the output sequence; that token gets passed as the input token for the next time-step. This step is repeated until an end-of-sequence token is decoded. Seq2seq models have been widely used in generative NLP tasks including machine translation \cite{sutskever2014sequence} and summarization \cite{nallapati2016abs}.


\subsection{Word Embeddings}

Word embeddings map discrete tokens like words into a real-valued vector space, taking the word's semantics into account. Some of the earliest approaches represented words as one-hot vectors, with a 1 at the index corresponding to the word. This na\"ive method had two major problems. First, it completely ignored semantics. As an example, \lq\lq Paris\rq\rq\space would be just as close to \lq\lq armchair\rq\rq\space in representation space as it would to \lq\lq France.\rq\rq\space Secondly, it was not scaleable, as it represented each word as a vector over the entire vocabulary. 

A very influential method using neural networks to create dense semantic word embeddings was word2vec \cite{mikolov2013distributed}. Word2vec introduced the skip-gram model, which learned word representations by predicting the surrounding context of a "center" word. It also introduced the continuous bag-of-words (CBOW) model, which predicted a "center" word given its context. Both produced meaningful word embeddings that proved excellent at making analogies. Later, Facebook AI proposed FastText \cite{joulin2017bag}, which modified word2vec by inputting n-gram character strings rather than full words. This minor adjustment significantly sped up training. GloVe \cite{pennington2014glove}, is a similar word embedding model. It learns embeddings from global statistical information on the number of times words co-occur together. Finally, following a similar approach to word2vec, doc2vec \cite{quoc2014doc2vec} is a model to learn fixed-length dense embeddings of sentences, paragraphs and documents.

\subsection{Language Models}

Word2vec, GloVe, and FastText embeddings produce fixed embeddings for a given word. Such a framework is limiting, however, because words may have many different meanings depending on their context. Some more recent models use RNNs to create such contextualized embeddings. Most of these are language models -- models that predict the probability of an input sequence by factoring it as product of conditional probabilities. Each conditional probability is the probability of a word given the entire sequence that precedes it. Language models make excellent embedding models because they learn to represent words based on prior context. 

An important model that used a language model to create word embeddings was ELMo \cite{peters2018deep}. ELMo, which stands for Embeddings from Language Models, introduced the idea of using bidirectional LSTMs to consider a word's context in both directions. That way, each embedding factors in every other word in the input, not just the preceding ones.


A significant breakthrough for language models is the Transformer model \cite{vaswani2017attention}. 
The Transformer is an encoder-decoder model that uses a modified attention mechanism called \lq\lq self-attention\rq\rq\space to represent text in a parallel fashion. Its novel network architecture trains faster and yields better results than traditional seq2seq models with attention. Rather than using a sequential RNN to encode the input one token after another, the Transformer processes each input token at the same time. This parallelization dramatically speeds up training. But it does not work in isolation; instead, the self-attention mechanism examines the representations of all other input tokens when encoding an input. The overall architecture is a stack of Transformer encoder layers, the final result of which gets passed to a stack of Transformer decoder layers. As with standard seq2seq models, the Transformer decodes the output one token at a time. 

Several models based on the Transformer have transformed the NLP landscape in the years following its release. OpenAI's Generative Pretrained Transformer (GPT) \cite{radford2018improving} was the first model to create word embeddings with Transformers. GPT, along with its subsequent versions GPT-2 and GPT-3, is a stack of Transformer decoders. All three work as language models, as the decoder outputs one successive token given all previous decoded tokens at each time step. With this framework, they are able to pretrain on vast amounts of unstructured text in an unsupervised manner. Then, once pretrained, the GPT models can be adapted for a variety of tasks such as question answering and summarization.

Bidirectional Encoder Representations from Transformers \cite{devlin2019bert}, or BERT, is another major breakthrough model based on the Transformer model.  It combines the benefits of bidirectional training found in the ELMo model \cite{peters2018deep} with the Transformer architecture, producing stellar results on dozens of NLP tasks. BERT, like ELMo and the OpenAI GPTs, is a language model that is pretrained on raw text and can be fine-tuned for many different tasks. However, unlike a standard language model, which predicts a word given the sequence of preceding words, BERT masks input words at random and trains to predict the masked words. The architecture of this \lq\lq masked language model\rq\rq\space is merely a stack of Transformer encoders. At the time of its publication in late 2018, it had already achieved state-of-the-art performance on 11 NLP tasks.

Several important variants of BERT followed that are also commonly used in NLP work. RoBERTa \cite{liu2019roberta} is a model that uses the same architecture as BERT but with adjustments that improve performance. For example, RoBERTa trains on more data for a longer period of time. It also dynamically adjusts the masking scheme and trains on longer sequences. T5 \cite{raffel2019exploring} and BART \cite{lewis2020bart} are models that adapt BERT to be better suited for text generation. Both of them add a Transformer decoder to the BERT encoder, decoding the original input one word at a time. They also have a more elaborate masking mechanism that includes blanks and masks of longer spans of text. Some other Transformer/BERT variation models focus on dealing with long documents, such as Transformer-XL \cite{dai2019transformer} and Longformer \cite{beltagy2020longformer}. In practice, clinical notes that contain thousands of tokens may take advantage of these models.

\subsection{Transfer Learning} 

Transfer learning has more recently shown immense usefulness in NLP \cite{mou2016transferable, ruder2019transfer,ruder2019neural}. Language models in particular are often used for transfer learning via pretraining. Their importance can be attributed to two causes: one, they train to develop an intuitive understanding of semantics and grammar, and two, they are unsupervised models that train on raw text data--a resource that is in wide supply. Models can pretrain on gigantic corpora like the entirety of English Wikipedia. Once pretrained, network architectures for language models like ELMo and BERT can be easily adapted for classification tasks like spam detection, sequence tagging tasks like named entity recognition, generative tasks like abstractive summarization, and many others \cite{mulyar2020mtclinical,peters2018deep}. The recent success of many pretrained models over existing benchmarks cements the status of transfer learning as an indispensable tool in contemporary NLP. In particular, EHR data can be scarce at times due to restrictions involving privacy and annotation availability, making transfer learning a frequently employed and important technique within the space of NLP for textual EHR data.

%% file: 2_classification.tex
\section{Classification and Prediction}

Text classification and prediction tasks in EHRs are critical to quickly processing thousands of large texts for clinical decision support, research, and process optimization. In this section, we cover several main subtasks, including general medical text classification, segmentation, and word sense disambiguation, 
as well as selected specialized subtasks for EHRs such as medical coding, outcome prediction, and de-identification.

\subsection{Medical Text Classification}

Significant prior work has been done in the text classification for medical texts and clinical notes. These works rely on traditional and classical methods like support vector machines (SVMs) and k-nearest neighbor (kNN). Marafino et al. \cite{marafino2014ngram} developed an SVM classifier that was able to identify a range of diagnoses and procedures in the intensive care unit (ICU). Including n-gram features was found to further improve performance. In Khachidze et al. \cite{khachidze2016natural}, SVMs and kNNs were comparatively applied to classify small instrumental diagnostics records in Georgian. After de-identification and simple preprocessing like tokenization and lemmatization, they performed feature selection and applied the classifiers. Results show that among three classes Ultrasonography, X-ray and Endoscopy, SVM outperform kNNs by around 4-5\% on the F1 score and 2-4\% on the accuracy. 


\begin{figure}
\centering
\includegraphics[width=\linewidth]{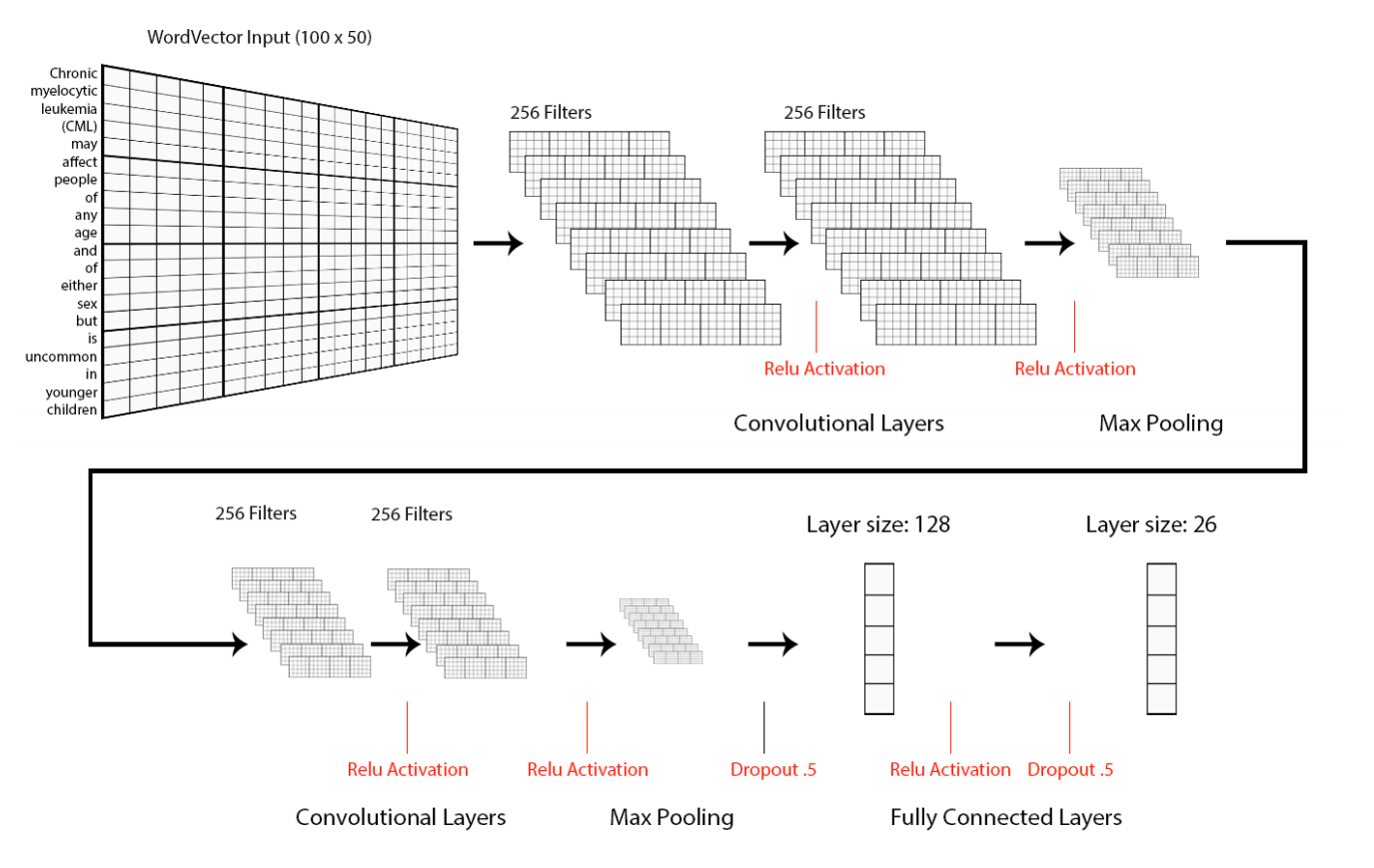}
\caption{Medical text classification using CNN. Adapted from the original paper. \cite{hughes2017medical}}
\label{fig:medical_cnn_hughes}
\end{figure}

In recent years, deep models like CNNs have attracted attention and achieved very competitive results on classification tasks. One of the early attempts was by Hughes et al. \cite{hughes2017medical}, who applied a CNN to classify clinical text at the sentence level. The model structure has four convolutional layers after the sentence embedding input, and at the end, a fully-connected layer is applied to predict the sentence labels. They compared their method with a variety of traditional machine learning methods and various sentence embeddings, including logistic regression, doc2vec embeddings \cite{quoc2014doc2vec}, and bag-of-words features. Results from Hughes et al. and others have proved that deep models including word embeddings and CNN models are competitive with and can actually outperform TF-IDF (Term Frequency - Inverse Document Frequency) and topic modeling features \cite{wang2019clinical,hughes2017medical}. In a similar way, Yao et al. \cite{yao2018clinical} introduced an approach to combine rule-based features and knowledge-guided deep learning techniques for supervised classification of diseases using clinical texts. When  labeled data for supervised learning is insufficient, it is also possible to use weak supervision with pre-trained word embeddings \cite{wang2019clinical}. Applying pre-trained models like BERT and BioBERT \cite{lee2020biobert} for classification is another alternative \cite{mascio2020comparative}.


Besides the previously mentioned generalized medical text classification tasks, other work focuses on specific applications in the medical domain. For example, recent works have been focused on identifying the chief complaint, or motivation, for medical visits. The chief complaint is about the patient’s medical history, current symptoms, and reason for visit. 
It is often encoded within the EHR in the form of free-text descriptions. 
This is essential for the development of automatic patient triage systems, and the data itself is helpful in various predictive tasks. The model proposed by Valmianski et al. \cite{valmianski2019evaluating} attempts a one-vs-rest classification of chief complaints in the EHR, comparing BERT-based models with a TF-IDF baseline model. Although the TF-IDF model outperformed the BERT-based models here, the authors suspect that this result is in part due to the short length of the median text entry; the TF-IDF model robustness remains questionable. Chang et al. \cite{chang2020generating} have also worked on applying a pre-trained BERT model to chief complaint extraction. They derived contextual embeddings to predict provider-assigned labels, focusing more specifically on emergency department chief complaints. They evaluated LSTM, ELMo and BERT and found that LSTM and ELMo achieved nearly the same performance with an average accuracy of 0.82 and 0.822, while BERT achieved 0.844.

Sepsis detection is another important classification task. Sepsis is a clinical condition defined as life-threatening organ dysfunction caused by infection \cite{mervyn2016third}. The patient's outcome is critically dependent on early detection and intervention, but symptoms are often missed because of shared similarities with other conditions. The requisite data for such diagnoses and predictions already exist in patient EHRs, so Futoma et al. \cite{futoma2017learning} developed an RNN classifier to detect sepsis in EHRs. They framed the problem as a multivariate time series classification problem. By using a multitask Gaussian process (MGP) and feeding into an RNN, they were able to make substantial improvements over baselines and clinical benchmarks with significantly higher precision. Specifically, compared with vanilla RNN, the model improved by 0.1 on the precision; compared with non-deep learning methods, the model improved the precision by 0.3-0.4.



\subsection{Segmentation}

Segmentation is the task of finding boundaries between sections of a text, and it serves an important role in preprocessing documents. The application of automatic segmentation in clinical documents is vital, as most EHRs of any length are either explicitly or implicitly segmented. In EHRs, topic segmentation aims to segment a document into topically similar parts; it is also known as text segmentation or discourse segmentation. Typically, text segmentation is formulated as a sentence classification task by either classifying each sentence into a correct category \cite{apostolova2009automatic}, or deciding if the evaluated sentence is a border sentence of its section \cite{badjatiya2018attention}.



There have been many attempts to use traditional machine learning methods to perform segmentation. One work by Apostolova et al. \cite{apostolova2009automatic} proposed a model for segmentation of biomedical documents. Using a hand-crafted training dataset of segmented clinical notes, they proposed an SVM to classify each sentence into a section, taking formatting and contextual features into account. 
Li et al. \cite{li2010section} also tried to take advantage of the format, noting that sections of a clinical note tended to follow a similar sequential pattern, and used a Hidden Markov Model (HMM) on a labeled dataset of clinical notes to infer the section labels. 
Later, Tepper et al. \cite{tepper2012statistical}, targeting the same formatting patterns, instead favored identifying likelier EHR section sequences by using beam search and a maximum entropy approach.




\begin{figure}
\centering
\includegraphics[width=\linewidth]{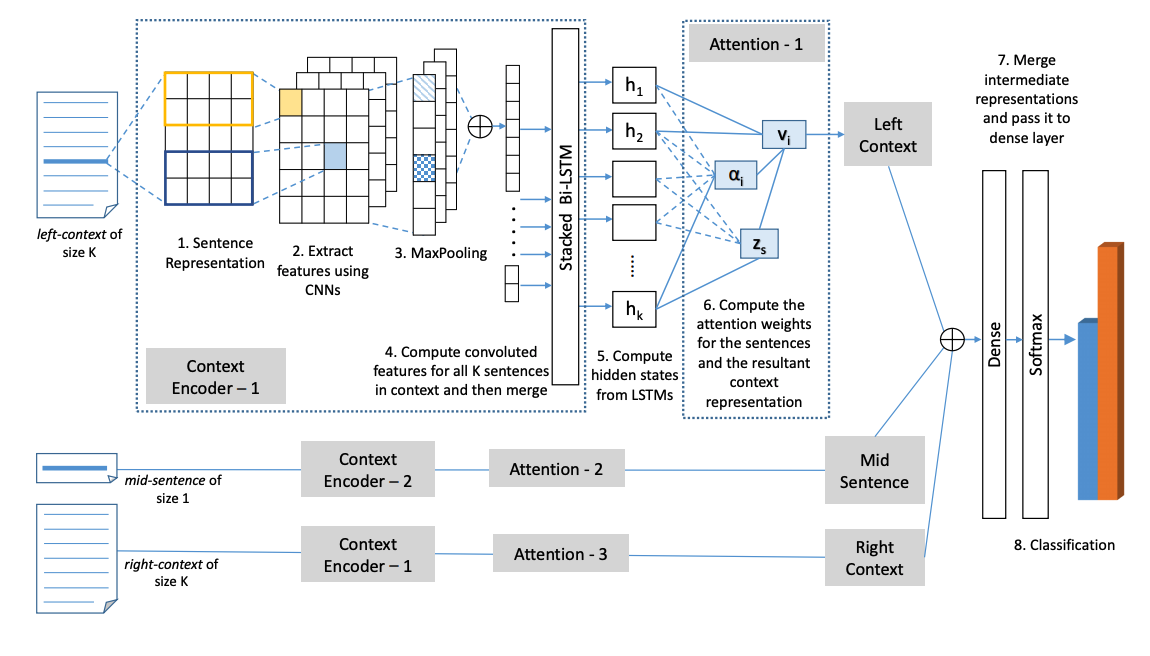}
\caption{Architecture diagram for the neural attention segmentation model. \cite{badjatiya2018attention}}
\label{fig:badjatiya2018attention}
\end{figure}

Recently, a few works have attempted to utilize deep models for segmenting medical texts. Word embeddings have been investigated to improve the segmentation for medical textbook chapters \cite{karususing,badjatiya2018attention}. Jatiya et al. \cite{badjatiya2018attention} applied attention mechanism for a neural segmentation model. They formulated segmentation as a binary classification task to decide if a sentence in the text is the beginning of a section. For each sentence, they used CNNs to create sentence embeddings for both the sentence in question and its context sentences on each side. The middle sentence attends to these sentence embeddings to create context vectors, and these representations are then merged and used for binary classification.


\subsection{Word Sense Disambiguation}

Word Sense Disambiguation (WSD) is used to assign the correct meaning to an ambiguous word given its context. In the medical domain, EHRs often contain many ambiguous terms that require specific domain knowledge. For example, the word \lq\lq ice\rq\rq\space may refer to frozen water, methamphetamine (an addictive substance), or caspase-1 (a type of enzyme) \cite{wang2018interactive}.

The WSD task has been approached with supervised learning, semi-supervised learning and knowledge-driven methods \cite{liu2004multi,xu2015clinical,finley2016towards}. These approaches show that massive high-quality annotated training data are essential to achieve a desirable WSD system performance. This is especially true for medical WSD, where only experts with substantial background knowledge can provide high-quality annotations. In addition to improving the clarity of the text, WSD can also help in downstream tasks like machine translation, information extraction, and question answering \cite{ramakrishnan2003question,chan2007word,zhong2012word}.

Some early works have applied machine learning models like  SVMs, Na\"ive Bayes, and decision trees to tackle WSD \cite{xu2006machine,bruce1994word,lee2002empirical}. There exist of attempts for applying neural methods for WSD. The deepBioWSD model \cite{pesaranghader2019deepbiowsd} is a representative work. First, they utilized the Unified Medical Language System (UMLS) sense embeddings; these embeddings are then applied to initialize a single bidirectional long short-term memory network (BiLSTM), which is then trained to do sense prediction for any ambiguous term.  Among some selected baseline models including LSTM, SVM and BiLSTM, the deepBioWSD outperformed in biomedical text WSD and achieved 96.82\% for macro accuracy, while the others are below or around 95\%. 
Other works investigated similar neural network structures, like multi-layer LSTMs \cite{bis2018layered},  BiLSTMs with self-attention \cite{zhang2019biomedical} and more. Results show that with deeper or attention layers, these models can outperform basic neural network models by around 2-4\% on the average testing accuracy.


The task of medical term abbreviation disambiguation is a special case of WSD. This task can also assist other downstream NLP tasks like sentence classification, named entity recognition, and relation extraction \cite{jiang2011study}.
In clinical notes, it is quite common for physicians, nurses or doctors to apply medical term abbreviations to represent drug names, disease names and other words. Depending on the medical specialty and contents of the EHR, these abbreviations can have a wide range of possible choices \cite{xu2015clinical,joopudi2018convolutional}. For example, there exist at least 5 possible word senses for the term MR, including magnetic resonance, mitral regurgitation, mental retardation, medical record and the general English word Mister (Mr.) \cite{li2019neural}. Medical term abbreviation is getting increasingly important as meaningful and standardized notes is helpful for both patients and doctors \footnote{\url{https://www.opennotes.org/}}.

\begin{figure}
\centering
\includegraphics[width=\linewidth]{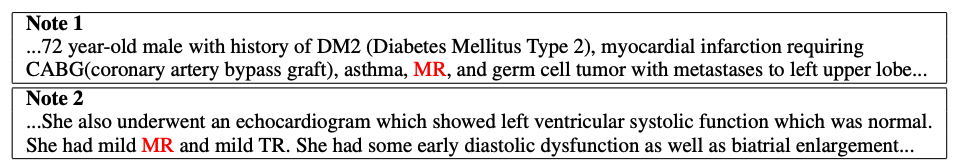}
\caption{An example for abbreviation disambiguation, adapted from the original work. \cite{li2019neural}}
\label{fig:abbreviation_example}
\end{figure}

\begin{figure}
\centering
\includegraphics[width=0.65\linewidth]{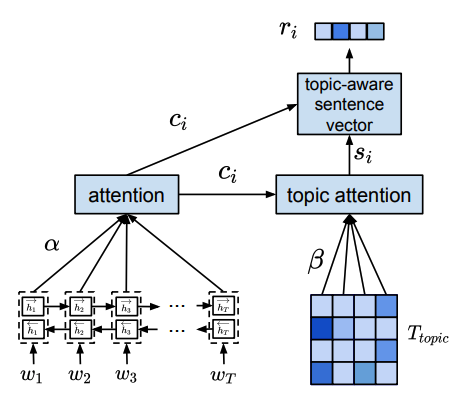}
\caption{The topic-attention model for abbreviation disambiguation. \cite{li2019neural}}
\label{fig:abbreviation_topic_atten}
\end{figure}

Some efforts have been made for abbreviation disambiguation on clinical notes. Traditional methods like decision trees were applied for acronym disambiguation in Spanish EHRs \cite{acronymDisSpanish}. Xu and Stetson \cite{xu2009methods} were the first to apply clustering techniques in building word sense inventories of abbreviations in clinical text. Later on, efforts were made to utilize word embeddings for abbreviation disambiguation. A work by Wu et al. \cite{wu2015clinical} examined three methods for word embeddings from unlabeled clinical corpus: surrounding based embedding, left-right surrounding based embedding and max surrounding based embedding. The word embeddings are treated as additional features to a WSD system which applies Support Vector Machines. 


Beyond word embeddings, more complex deep architectures have also been investigated. In a recent paper, Joopudi and Dandala \cite{joopudi2018convolutional} proposed a CNN model that encodes representations of clinical notes and predicts the meaning of an abbreviation using classification techniques. Moreover, a neural topic-attention model was applied to learn improved contextualized sentence representations for medical term abbreviation disambiguation \cite{li2019neural}. In this work, a latent Dirichlet allocation (LDA) model was leveraged to learn topic embeddings, and then contextualized word embeddings (ELMo) were applied to conduct a topic-aware sentence vector for classification. Adams et al. \cite{Adams2020zeroclinical} introduced another deep latent variable model, called Latent Meaning Cells (LMC), for clinical acronym expansion, focusing on utilizing local context and document meta-data for contextualized representation.

\subsection{Medical Coding}

The medical coding task attempts to map text from EHR to International Classification of Diseases (ICD) codes \cite{scheurwegs2015data}. These codes represent different diagnoses, and they are used in clinical treatment, medical billing, and statistics collections. ICD codes can help physicians quickly determine which diseases are involved and reach clinical decisions in a more timely manner, but are also tediously specific and detailed. For example, diabetes alone has over two dozen different codes. Human coders struggle to manage the scale and complexity of the processes and often make costly mistakes. Many attempts have been made to improve ICD coding using rule-based methods \cite{Farkas2008} and machine learning algorithms like Bayesian Ridge Regression \cite{lita-etal-2008-large}, SVMs \cite{perotte2014diagnosis,Koopman2015}, and more.


\begin{figure}
\centering
\includegraphics[width=\linewidth]{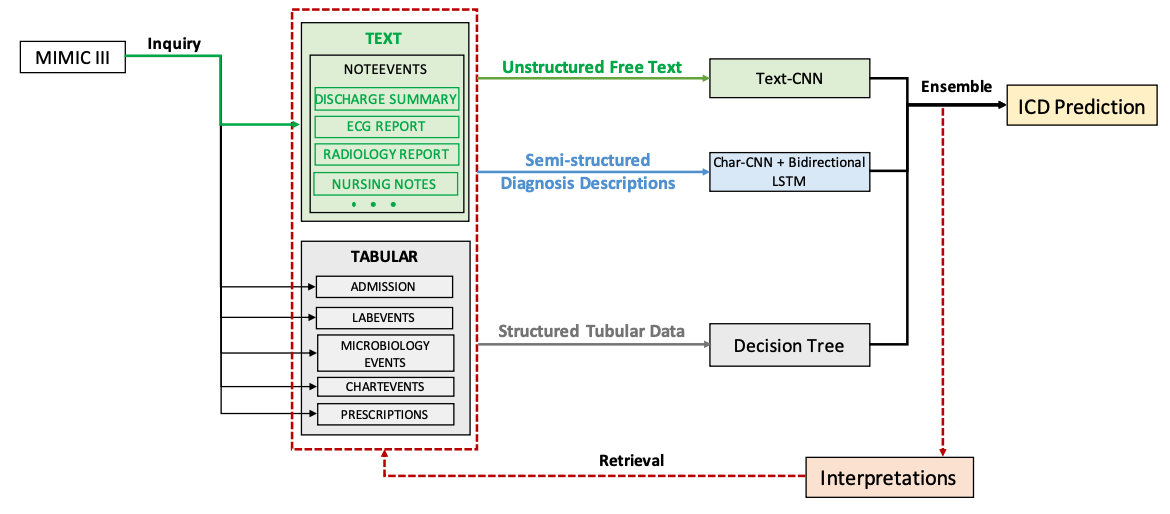}
\caption{A multimodal model architecture for ICD code prediction. \cite{pmlr-v106-xu19a} }
\label{fig:icd}
\end{figure}

Many earlier deep learning-based methods have focused on applying CNNs, LSTMs and LSTMs with attention by formulating this task as a document classification task \cite{pmlr-v106-xu19a,shi2017automated,tal2017multi}. A representative model with explainability is Convolutional Attention for Multi-Label classification (CAML) by Mullenbach et al. \cite{mullenbach2018medical}. It is a CNN-based model with attention. The model first aggregates information from the entire document using a CNN. After that, attention mechanism is applied to select the most relevant segments from the document which trigger ICD code prediction, providing explanations in the process. Xu et al. \cite{pmlr-v106-xu19a} proposed a multimodal framework which considers unstructured texts, semi-structured texts, and tabular data when predicting an ICD code. A CNN is applied for modeling the unstructured texts. Then a deep learning model which contains a character level-CNN and a BiLSTM is applied for the semi-structured text. Finally a decision tree is applied for the tabular data. By assembling these models, the system is able to make ICD code predictions. Focusing on features extracted from discharge summary notes in MIMIC-III, a widely used, albeit limited, EHR dataset, Huang et al. \cite{Huang_2019} conduct a study on multi-label ICD-9 code classification. They tested 1) a CNN-based classification model with word2vec document representations for each discharge summary, 2) a standard LSTM-based model, and 3) a GRU-based model on the sequential discharge summary text. They conducted two main classification tasks: predicting the top 10 and top 50 of ICD-9 codes. First, on the top 10 codes, compared to several baseline methods, including logistic regression, random forest, and feed-forward networks, RNN-based methods showed greater performance in prediction. However, on the top 50 codes, logistic regression outperformed the proposed deep method, leaving room for improvement in deep-learning methods for multi-label ICD-9 coding.

Recent improvements in pre-trained language models have greatly helped automatic medical coding. Singh et al. \cite{singh2020multilabel} implemented a BERT model to predict ICD-9 codes from unstructured clinical text as a multi-label classification task. They pre-trained the BERT model the MIMIC-III dataset, allowing this model to learn the medical data context with less computation and preprocessing than other methods. Results show that this model significantly out-performs existing literature, performing better as the classifier handles more diagnosis and procedure codes.

Medical coding suffers from a severe data imbalance between the most and least popular codes because some codes are used much more frequently than others. Vu et al. \cite{Vu_2020} attempt to solve this by proposing an attention-based model, which is able to adapt to the varied interdependence and lengths of text fragments. The labelling attention model consists of four layers: a pre-trained embedding layer, a bidirectional LSTM, an attention layer for label-specific weight vectors, and label-specific binary classifiers. They then extended the label attention model into a hierarchical joint learning mechanism called JointLAAT to handle the data imbalance for rare codes. 

Another challenge in medical coding is capturing higher-level information from clinical encounters. Each encounter can have multiple associated documents, and coding is often done at this level rather than on individual documents. Shing et al. \cite{shing2019assigning} proposed an encounter-level document attention network (ELDAN), which was made of three parts: a document-level encoder that turns sparse document features into dense document features, a document-level attention layer, and an encounter-level encoder. They treated the problem as multiple one-vs-all binary classification problems, and the model outperforms the baseline without the need to train on document-level annotations.


\subsection{Medical Outcome Prediction}

Prediction of medical outcomes (e.g. length of stay, progression to heart failure, death) is challenging but existing available information such as providers notes and imaging reports can help.  
We highlight representative deep learning approaches that utilize structured EHR data for this task, then introduce how unstructured text could help.

Doctor AI \cite{choi2016doctor} is an early attempt that predicts new diagnoses and medications for a patient's subsequent visit. The researchers used an RNN because its sequential architecture lends itself to modeling the temporal nature of a patient's healthcare trajectory. The input at each time-step is a raw representation of a single patient visit, incorporating relevant diseases, medication, and procedure codes. The hidden state serves as a representation of the patient's medical history at that point in time.
In a similar temporally-based prediction task, Suresh et al. \cite{suresh17a} used LSTM and CNN-based models for forward-facing predictions of ICU intervention tasks, including ventilation, using vasopressors, and using fluid boluses. The data is split into 6 hour chunks, where patient data is recorded, as well as the status of the interventions being taken. After a 6 hour gap period, a 4 hour prediction period is allocated to test the model and predict which interventions were taken during this period. They report high AUC (area under curve) scores on LSTM and CNN-based models compared to a linear regression base model, with an improvement of to 0.24 AUC.

Electronic health records tend to have irregular intervals between observations, as clinical events are often sudden occurrences. DeepCare \cite{pham2017predicting} is an end-to-end neural network that addresses the episodic nature and irregularity of EHRs, validated on mental health and diabetes datasets. It reads medical records, stores illness history, infers current health status and predicts future medical outcomes. Each care episode is represented by a vector; the relationship between them is modeled with a modified LSTM that accounts for irregularities in time, admission methods, diagnoses, and interventions. DeepCare aggregates the patient's medical history using multi-scale temporal pooling, then predicts the probability of specific medical outcomes.
Lyu et al. \cite{lyu2018improving} further explored medical time series, choosing to investigate unsupervised representation learning methods. and A seq2seq model was applied as a forecaster. They found that the forecasting seq2seq model performed best, and particularly that an integrated attention mechanism can improve clinical predictions, with a mean squared error of 0.098 with attention compared to 0.119 without. 

Recently, Zhang et al. \cite{MetaPred:19} proposed MetaPred, a transfer-learning framework to assess clinical risk for low-resource clinical disease data, to address the limited number of sets of labeled data samples. MetaPred was trained on related risk prediction tasks to learn how good predictors are learned, and this meta-learned model can be fine-tuned or applied on the target disease data. With CNN and RNN as base predictors, MetaPred outperformed fully supervised models in classifying mild cognitive impairment, Alzheimer's, and Parkinson's as measured by AUCROC and F1 score. On average, MetaPred improves about 0.3464 on AUCROC and 0.4521 on F1.
		
In a task closely related to outcome prediction, Hsu et al. \cite{hsu2020characterizing} addressed the predictive power of the variety of unstructured notes within EHRs. Because some unstructured notes include heavy copying from the structured fields, they investigated how important unstructured notes are for prediction tasks, and evaluated which parts of notes are the most important. Using MIMIC-III, they considered readmission prediction and in-hospital mortality prediction to determine the value of unstructured notes. They found that unstructured notes are useful in select situations (e.g. readmission), but provide negligible additional value in others (e.g. mortality prediction). However, trained a small group of well-selected sentences, as chosen by value functions considering length, fractions of medical terms, etc., the model outperformed the entire set of notes for downstream prediction tasks.

\subsection{De-identification}
\label{sec:de-identification}

De-identification is the task of removing patient sensitive information and preserving clinical meaning in EHR data. De-identification of PHI (protected healthcare information) in EHR clinical notes is a critical step to protect patient privacy before sharing or publishing datasets for secondary research purposes. The Health Insurance Portability and Accountability Act (HIPAA) includes 18 different types of PHI \footnote{\url{https://www.hhs.gov/hipaa/index.html}} including names, locations, and phone numbers. 
Several shared tasks have been organized to promote de-identification of clinical text in the NLP community \cite{stubbs2015automated,uzuner2007evaluating,stubbs2017identification}. Given the large amount of EHR data needed, manual de-identification is not feasible because of the amount of time it requires. Rule-based de-identification studies mainly depend on dictionary pattern matching and regular expressions \cite{stubbs2015automated}, but these methods usually require complex algorithms and cannot handle unexpected cases such as typos or infrequent abbreviations. 

Today, it is common to utilize automatic de-identification approaches that apply named entity recognition (NER) methods. Some other automatic de-identification tools use hybrid methods which combine rule-based methods and machine learning based NER methods \cite{stubbs2015automated} to achieve good performance, reaching as high as 0.9360 micro-averaged F1 on the 2014 i2b2/UTHealth de-identification shared task. However, these methods still have difficulties when applied to data in languages other than English, or when used in different clinical domains.

Early de-identification systems for patient records based on neural networks include LSTMs or LSTM-CRF (conditional random field) models \cite{dernoncourt2017identification, yang2019study}. These systems consisted mainly of four components: an embedding layer, a label prediction bidirectional LSTM layer, a CRF layer and a label-sequence optimization layer. They marginally outperformed CRF-based approaches (i.e., improves 0.01 on the F1 score), demonstrating that the CRF model is still a competitive baseline.
 
   
Some studies additionally found that the embedding layer has a large impact on the model performance. Wu et al. \cite{wu2018combine} showed that an RNN model integrated with medical knowledge from UMLS outperformed a baseline RNN model. A study by Tang et al. \cite{Tang2020deid} compared several embeddings, including common models like CBOW, Skip-Gram\cite{mikolov2013distributed}, and GloVe\cite{pennington2014glove}, and contextual embeddings like BERT and ELMo, using the representative method of BiLSTM-CRF. The BERT embedding had better overall performance than any other evaluated embedding method.

While previous works focused on English texts, there have been many efforts toward de-identification in non-English medical texts, where datasets are even more limited. 
Kajiyama et al. \cite{kajiyama-etal-2018-de} applied rule-based, CRF, and LSTM-based methods to three Japanese EHR datasets. 
They observed that LSTM outperformed CRF by 40\% on F1 score on the MedNLP-1 dataset (Japanese EHRs), and about 7\% on a mixture of dummy EHR and MedNLP-1 dataset, in which the dummy EHR is built by themselves and contains EHR of 32 hospitalized patients. 
Trienes et al. \cite{trienes2020comparing} compared a series of de-identification methods, including a rule-based system, a feature-based CRF, and a deep neural network (BiLSTM-CRF), for transferability and generalizability from English to Dutch, and found that the deep neural network performed best. 
García-Pablos et al. \cite{garcapablos2020sensitive} tested a pre-trained multilingual BERT model on several Spanish clinical texts. They compared the BERT-based sequence labelling model with a baseline sensitive data classifier, the spaCy Spanish NER model, and CRFs. They showed that the simple BERT-based model without domain-specific fine-tuning was able to out-perform all other methods and was also robust to training data scarcity. These works show that applying pre-trained models is also becoming a trend for non-English tasks, due to the capacity of multilingual BERT models. 

%% file: 3_clinical_embeddings.tex
\section{Embeddings}

This section explores various applications of clinical embeddings in the EHR domain. Such representations have been used to model the semantics of biomedical text, the health trajectory of individual patients, and many other tasks. We also describe how recent pre-trained language models can promote better representation learning methods in EHR and biomedical text.

\subsection{Medical Concept Embeddings}

Medical concepts include various types of entities: genes, proteins,  diseases and more. Learning the semantics of these medical concepts can be helpful for other applications. A number of papers use deep models to learn such domain-specific embeddings. 

A representative work, cui2vec \cite{beam2020clinical} was trained on a corpus of 80 million documents and over 100k medical concepts including insurance claims, clinical notes, and journal articles.
As a preprocessing step, the researchers mapped each medical concept word or phrase to its corresponding \lq\lq concept unique identifier,\rq\rq\space or CUI. The co-occurrence statistics from these mappings are used to construct concept embeddings with GloVe \cite{pennington2014glove} and word2vec \cite{mikolov2013distributed}.




Many concepts, including lab tests, diagnoses and drug administrations, are temporal in nature. These \lq\lq heterogeneous temporal events\rq\rq\space may have a high degree of correlation between them. For example, the event of some diagnosis being made is correlated with the results of certain lab tests. In addition to understanding the relationships between events, the model must also learn temporal representations. Some medical events happen only once, whereas others will occur periodically according to some treatment patterns. Ignoring the dynamics of the embedded units may lead to poor semantics when learning representations. Liu et al. \cite{liu2018learning} proposed a model for obtaining the joint representation of heterogeneous temporal events. The model is trained on the overarching classification task of clinical endpoint prediction, which predicts whether some medical event like a disease or symptom will happen in the future. Their main contribution is a modified LSTM cell based on the Phased LSTM model proposed by Neil et al. \cite{neil2016phased}. The Phased LSTM alters the traditional LSTM by adding a time gate that accounts for inputs with irregular sampling patterns. Liu's model goes a step further by adding an event gate that is capable of modeling correlations among thousands of event types. Zhu et al. \cite{zhu2019measuring} also preserve the temporal properties by compressing each patient visit into a fixed-length vector with medical embeddings based on surrounding medical context. These event embedding vectors are stacked together to produce a dense embedding matrix for each patient. Pairs of these matrices are passed through convolutional filters and mapped to feature maps, which are then pooled into intermediate vectors to build the patient representation embeddings. A symmetrical similarity matrix is constructed using the distance between the feature vectors. Their proposed framework achieves strong performance on similarity measures, with an R1 value of 0.99 and Purity of 0.99 as opposed to baseline approaches such as KMeans clustering, with an R1 value of 0.66 and Purity of 0.42.

\subsection{Visit Embeddings}

Visit embeddings of patients are also crucial for clinical decision-making, outcome prediction, and other downstream tasks such as question answering and forecasting events.

Some methods treat patient visits as time-series data. An early attempt, Med2vec \cite{choi2016multilayer}, generates vector representations of patient visits and medical concepts like procedures, diseases, and medications. It represents each patient visit as a vector of medical codes corresponding to the concepts that occurred in the visit. Like the word2vec skip-gram model, med2vec reads the current visit, embeds it, and predicts the likelihood of previous and future visits. Doing so preserves the sequential order of a patient's visits and the co-occurrence between concepts. 
Med2vec embeddings outperform previous models at tasks like predicting concepts in future visits and calculating the patient’s current severity status. 

However, most models represent each patient visit as a flattened collection of concepts like diagnoses and treatments. Doing so, however, ignores valuable hierarchical information on the multilevel relationship between the concepts in an EHR. For example, the diagnosis of a fever can lead to treatments like acetaminophen and IV fluids, which can in turn produce side effects that need treatments of their own. MiME \cite{choi2018mime} models the inherent hierarchical structure of EHRs, leveraging it to create embeddings at the concept, visit, and patient level while jointly performing auxiliary prediction tasks. These multilevel embeddings can predict heart failure and sequential diseases with low test loss (0.25), as opposed to more traditional NN activation functions, with linear activation at 0.26 and tanh likewise at 0.26. 

The neural clinical decision support system by Wei et al. \cite{wei2018embedding} created visit representations in a simple yet clever way. First, they extract diagnostic ICD codes from the MIMIC-III database. With these as labels, they train a CNN to predict the patients’ codes from the raw EHR text. The final dense layer of the network is taken to be a visit representation. This representation is successfully applied to an information retrieval task of recommending relevant literature for individual patients, achieving a MAP score of 0.2084 when combined with cosine similarity. Here, visit representations from MIMIC help achieve strong performance on a task for which little training data exists. 
The deep neural network by Escudié et al. \cite{escudie2018deep} learns low-dimensional representations of patient visits from which ICD codes have been removed to predict the presence or absence of such codes. A CNN applied to the text features and a multi-layer perceptron (MLP) used on the structured data are trained together. The last hidden layers of each sub-network are concatenated to produce an embedding for each stay. This embedding is able to conserve semantic medical representation of the initial data and improves the prediction performance of a random forest classifier, increasing the average F1 score among predicted codes from 0.595 for the random forest to 0.754 with the embeddings, and 0.820 with the artificial neural network.

\subsection{Patient Embeddings}

Patient embeddings are another way to take advantage of EHR understanding and secondary usage and enable better decision-making and outcome predicting.

As an early attempt, Mehrabi et al. \cite{mehrabi2015temporal} proposed a deep learning method for temporal pattern discovery over the Rochester Epidemiology Project dataset by modeling individual patient records as a matrix of temporal clinical events. The rows of the matrix represent medical codes, and the columns represent years. The embeddings are created from this matrix with an unsupervised network called a deep Boltzmann machine. This model is highly restricted, however, because it only utilizes 70 most frequent ICD-9 codes. In addition, all temporal information is chunked in units of one year, which may be too long in the medical context.

\begin{figure}
\centering
\includegraphics[width=\linewidth]{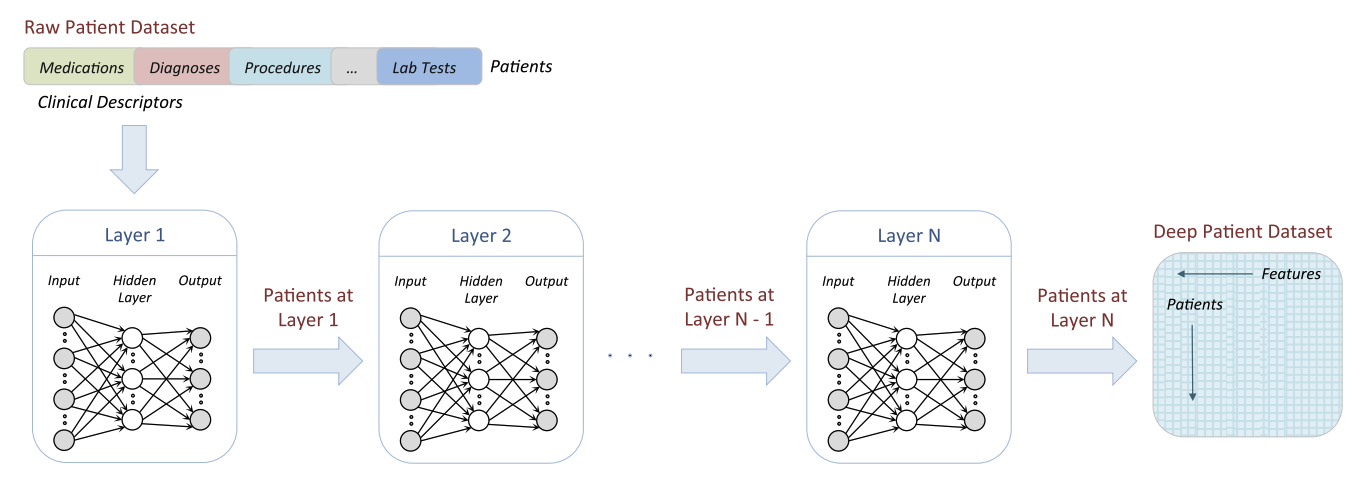}
\caption{The diagram of the Deep Patient framework, adapted from original paper\cite{Miotto2016}: an unsupervised deep learning method which is trained on raw dataset, and is able to learn patient representation in the last neural network layer. }
\label{fig:deeppatient}
\end{figure}

A more widely-used model, Deep Patient \cite{Miotto2016}, is a deep learning framework for general-purpose patient representation learning from EHR data. It extracts raw features like ICD-9 codes, medications, lab tests, and concepts from preprocessed EHRs. Each patient is represented with either a single vector or by a sequence of vectors determined by temporal windows. The model then embeds these raw vectors with stacked denoising autoencoders. These patient embeddings are applied to the task of clinical disease prediction. Evaluation was conducted on 76,214 test patients comprising 78 diseases from diverse clinical domains and temporal windows. 
Results show that this model improves accuracy and F-score by 15\% and 54\% respectively when compared with other baselines including K-means, PCA and so on.

Later, Dligach and Miller \cite{dligach2018learning} built upon methods from Deep Patient to learn patient representations using only text variables. The neural network model takes a set of CUIs as input and produces a vector representation of the patient. The final network layer is composed of Sigmoid units that are used to jointly predict all possible billing codes associated with the patient. The learned dense patient representations are successful in outperforming sparse patient representations on average and for most diseases. 

Similar to Deep Patient, the patient2vec model \cite{zhang2018patient2vec} represents each patient visit as a sequence of ICD-9 codes, medications, and lab tests. It learns an embedding for each of these codes by using word2vec to predict the codes that are likely to co-occur in a visit. Then it represents a patient’s entire history in a single embedding with an RNN and attention mechanism. Patient2vec predicted future hospitalizations with higher statistical power than previous patient embedding models, specifically, it achieves 0.02 gain on F2-score when compared with BiRNN-based models.

Unlike the systems described so far, Sushil et al. \cite{sushil2018patient} learned unsupervised patient representations directly from clinical text. They attempted this with two neural approaches: a stacked denoising autoencoder and a doc2vec model \cite{le2014distributed}. 
These neural networks produce patient representations that improved 0.03 upon doc2vec model on F1-score in  predicting mortality and primary diagnostic category. 

Finally, as an example which takes advantages of pre-trained models, TAPER \cite{darabi2020taper} uses text and medical codes to produce a unified representation from a patient’s visit data that can be used for downstream tasks. The medical code embedding is learned by a skip-gram model using the Transformer model, while a pretrained BERT \cite{devlin2019bert} model produces the medical text embeddings. These two embeddings are concatenated for the final patient representation. TAPER demonstrated about 5\% recall improvement upon med2vec when evaluated on code embedding using recall@k. Besides, it showed up to 67\% on the AUCROC score when applied on tasks of predicting readmission, mortality, and length of stay compared to baselines including med2vec and patient2vec.

\subsection{BERT-based Embeddings}


\begin{figure}
    \centering
    \includegraphics[width=\linewidth]{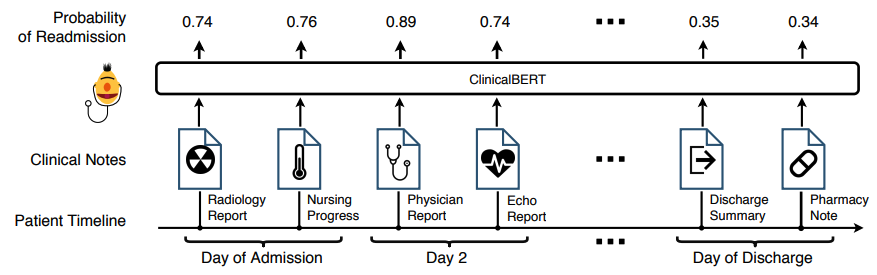}
    \caption{ClinicalBert illustration\cite{huang2019clinicalbert}: notes were added to electronic health record during a patient’s admission to update the patient’s risk of being readmitted within a 30-day
    window. }
    \label{fig:clinicalbert1}
\end{figure}

BERT \cite{devlin2019bert} harnesses the power of Transformers to generate better word embeddings than ever before. However, its embeddings generalize poorly to text from specific domains like biomedicine. For this reason, several works subject the pretrained BERT to a subsequent round of pretraining--this time, on corpora of bioinformatics or EHR text. 
BioBERT \cite{lee2020biobert} adapts BERT for biomedical domain. It is directly pre-trained on biomedical research papers from two large corpora:
PubMed abstracts (PubMed) and PubMed Central full-text articles
(PMC). Besides, it applies three representative biomedical text mining tasks for fine-tuning: named entity recognition, relation extraction and question answering.
SciBERT \cite{beltagy2019scibert} trains BERT on 1.14 million biomedical and computer science articles from the Semantic Scholar corpus. And, most relevantly to EHRs, ClinicalBERT \cite{huang2019clinicalbert} trains on clinical notes from the MIMIC-III corpus. 
EhrBERT \cite{li2019finetuning} is also trained on clinical notes, but it is not generally available because its training dataset is not public. Four datasets were included in the evaluation: the Medication, Indication, and Adverse Drug Events (MADE) 1.0 corpus, the National Center for Biotechnology Information (NCBI) disease corpus, and the Chemical-Disease Relations (CDR) corpus. Results show that EhrBERT outperforms other baseline systems including BioBERT and BERT by up to 3.5\% in F1.


MT-Clinical BERT \cite{mulyar2020mtclinical} is a special model that builds connections between individual tasks. In addition to learning embeddings of clinical text, it performs multitask learning on eight information extraction tasks including entity extraction and personal health indicator (PHI) identification. These embeddings are shared as inputs to these prediction tasks. This multitask system is competitive with task-specific information extraction models, as it shares information amongst disjointly annotated datasets.

BEHRT \cite{li2020behrt} is a deep neural transduction model that learns about patients’ past diseases and the relationships that exist between them. It uses the masked language model pretraining approach from BERT. Specifically, given the past EHR of a patient, the model is trained to predict the patient's future diagnoses (if any). BEHRT produces a final embedding that preserves the timing of events along with data concerning disease sequences and delivery of care. Evaluation of the model demonstrated BEHRT’s superior predictive power with an improvement of 8.0–13.2\% in average precision scores in tasks including disease trajectory and disease prediction, when compared to other approaches like RETAIN \cite{choi2016retain}.

MS-BERT \cite{Costa2020multiple} is a transformer model trained on real clinical data, rather than the MIMIC corpus. The model is trained on over 70,000 Multiple Sclerosis (MS) consult notes and is publicly available\footnote{\url{https://huggingface.co/NLP4H/ms_bert}}. Before training, the notes are de-identified. Then, the model is tested on a classification task to predict Expanded Disability Status Scale (EDSS), which is usually inside unstructured notes. The model surpasses other models that applied word2vec, CNN and rule-based methods on this task by up to 0.12 on Macro-F1 score.

CheXbert \cite{smit2020chexbert} applies BERT to the task of labeling free-text radiology reports. Existing machine learning methods in this task either employ feature engineering or manual annotations from experts. While of high quality, the annotations are sparse and expensive to create. CheXbert overcomes this limitation by learning to label radiology reports using both annotations and existing rule-based systems. It first trains to predict the outputs of a rule-based labeller, then fine-tunes on an augmented set of expert annotations. It set a new state of the art result by achieving an improvement of 0.007 on the F1 scores for a report labeling task on the MIMIC-CXR dataset \cite{Johnson2019MIMICCXRAL}, a large-scale labeled chest radiographs.



%% file: 4_extraction.tex
\section{Information Extraction}

Information extraction (IE) is the task of automatically identifying important content in unstructured natural language text. It encompasses several subtasks, including named entity recognition, event extraction, and relation extraction. In this section, we present an overview of various IE tasks and methods as applied to EHRs.



%


\subsection{Named Entity Recognition}

Named Entity Recognition (NER) is the task of determining whether tokens or spans in a text correspond to certain \lq\lq named entities\rq\rq\space of interest, such as medications and diseases \cite{leser2005named,gorinski2019named}.


Formulated as a sequence tagging task, NER is typically modeled using  Conditional Random Fields (CRFs) and/or RNN-based models \cite{huang2015bidirectional,gorinski2019named}. For example,
Cho et al. \cite{cho2019biomedical} proposed extracting biomedical entities by using a bidirectional long short-term memory network (BiLSTM) and a CRF, achieving improvements of up to 1.5\% on F-score over other methods, including BiLSTMs, and a Vanilla BERT. Similarly, Giorgi et al. \cite{giorgi2018transfer} combined both gold-standard and silver-standard corpora, and trained a LSTM-CRF model for biomedical NER. Du et al. \cite{du2019extracting} used a Transformer encoder and an LSTM decoder to extract symptoms within transcribed clinical conversations. In addition to identifying symptoms within the text, they also tried to predict whether or not the patient is experiencing the symptom. 

Recent works focus on transfer learning methods, including using pretrained word embeddings to solve medical NER tasks. Gligic et al. \cite{gilgic2020named} showed that medical NER performance can be improved by first pretraining word embeddings on unannotated EHRs with word2vec \cite{mikolov2013distributed}. They achieved about 0.95 F1 on the i2b2 Medical Extraction Challenge, improving several points past the previous state of the art. As far as embedding models go, however, BERT \cite{devlin2019bert}, which incorporates contextual information, is more sophisticated than word2vec. Because of this, the creators of BioBERT \cite{lee2020biobert} trained regular BERT on biomedical text, then fine-tuned it for a number of NER tasks. BioBERT's embeddings proved very effective at recognizing entities such as diseases, species, proteins, and adverse drug reactions. So, Yu et al. \cite{Yu2019biobert} evaluated a BioBERT-based NER system and showed that their system outperforms other models (i.e. BiLSTM-based and Attention-based ones) on three biomedical text mining tasks, achieving 0.62 additional F1 points on biomedical NER, and also improved on the other two subtasks.
Peng et al. \cite{peng2019transfer} similarly adapted BERT for biomedical text via pretraining on large biomedical datasets like PubMed and MIMIC-III; their version outperformed BioBERT on 10 different NER tasks like recognizing diseases, chemicals, and disorders.

While effective, these BERT models fail to address a central issue in medical NER: the transferability between medical specialties. Clinicians in different specialties use vastly different vocabularies, which poses a huge challenge in training an effective specialty-independent medical NER model. This problem is exacerbated by the scarcity of publicly available data, especially for certain specialties. Wang et al. \cite{wang2018labelaware} approached this issue by developing a double transfer learning framework for cross-specialty NER. Their system transfers both feature representations and parameters, which enables resource-poor specialties to utilize knowledge gleaned from specialties with more annotated EHRs.

As a sub-task of NER, clinical concept extraction seeks to identify medical concepts, such as treatments and drug names. In earlier years, challenges for concept extraction tasks were primarily won by hybrid approaches, which combined rule-based and machine learning-based approaches \cite{long2005extracting,khin2017medical}. Similar to NER, people have framed clinical concept extraction as a sequence tagging problem and commonly use LSTMs and/or CRF frameworks \cite{jagannatha2016bidirectional, chalapathy2016bidirectional,habibi2017deep}. For example, Ji et al. \cite{ji2020fully} proposed a multi-layer fully-connected LSTM-CRF model to conduct concept extraction, which cooperates with character-level word representations and pretrained word embeddings. Without any feature engineering or prior knowledge, the model outperforms other selected baseline models like CRF methods, with an F1 score of 0.845 on the i2b2 dataset.

Similarly, pretrained models and transfer learning techniques are also investigated for clinical concept extraction \cite{zhu2018clinical,neumann2019scispacy,lee2020biobert} .
A work by Tao et al. \cite{tao2018effective} embedded each word in an EHR note, then used the embeddings to predict whether they denote a medical concept. Each embedding is a concatenation of two separate representations: one derived from ELMo (which is an unsupervised bidirectional language model), and the other from numerous publicly available medical ontologies and lexicons, including the Wikidata graph \cite{vrandevcic2014wikidata}, the Disease Ontology \cite{schriml2012disease, kibbe2015disease}, and public FDA datasets. The latter enables the model's embeddings to draw from a larger domain-specific knowledge base. The model performs prediction with a CRF, which takes neighboring tokens into account when classifying each word. A recent system by Krishna et al. \cite{krishna2020extracting} extracted diagnoses and organ abnormalities from doctor-patient conversations. These conversations are often very long and contain large sections of clinically irrelevant information, so the researchers first filtered utterances by how "noteworthy" they are. From these utterances, they used a BERT-based model to recognize the diagnoses and abnormalities. Finally, Datta and Roberts \cite{surabhi2020hybrid} applied a BERT-based classifier pretrained on MIMIC-III data, and a filtering mechanism to identify spatial expressions in radiology reports. These expressions are important medical concepts that help describe radiographic images, and have use cases in image classification as well.

\subsection{Entity Linking}

Entity linking (or named entity linking) is the process of associating mentions of recognized entities to their corresponding node in a knowledge base. In practice, entity linking is helpful for automatic linking of EHRs to medical entities, supporting downstream tasks such as diagnosing, decision making and more \cite{jin2018improving}.

MEDTYPE \cite{vashishth2020medtype} presented a toolkit for medical entity linking by incorporating an entity disambiguation step to filter out unlikely candidate concepts. This step predicts the semantic type of an identified mention based on its context. They also utilized pretrained Transformer-based models as encoders.
Additionally, they introduced two large-scale datasets, WikiMed and PubMedDS, to bridge the gap of small-scale annotated training data for medical entity linking.

Zhu et al. \cite{zhu2020latent} introduced Latent Type Entity Linking model (LATTE), a neural network-based model for biomedical entity linking. In this case, latent type refers to the implicit attributes of the entity. The model consists of an embedding layer that contains semantic representations of the mentions and candidates. An attention-based mechanism is then used to rank candidate entities given a mentioned entity. This model outperforms previous ranking models, achieving over 0.92 on the MAP score for two entity linking datasets.


Oberhauser et al. \cite{oberhauser2020trainx} introduced TrainX, a system for medical entity linking that contains an named entity recognition system and a subsequent linking architecture. It is the first medical entity linking system that utilizes recent BERT models. They showed that TrainX could link against large-scale knowledge bases, with numerous named entities, and that it supports zero-shot cases where the system has never seen the correct linked entity before.



\subsection{Relation and Event Extraction}

The task of relation extraction identifies entities that are connected through a relation fitting specific relation types. This is critical in a health context because an NLP system must grasp the relationships between various medical entities in order to fully understand a patient's record. Some papers on relation extraction, especially early work \cite{Roberts_2008, rink2011, KIM2015274}, treat the task as a multi-label classification problem and train traditional classifiers, such as support vector machines, where the classes are typed relations. Since then, however, deep relation extraction systems have modeled more complex features of text, leading to improvement in modeling entities and their relations.

Several papers have used CNNs to model relation extraction. Sahu et al. \cite{sahu-etal-2016-relation} used a standard CNN augmented with word-level features to extract relations from clinical discharge summaries. Another work combined CNNs and RNNs to extract biomedical relations from linear and dependency graph representations of candidate sentences \cite{zhang_2018_hybrid}. This hybrid CNN-RNN model was benchmarked on various protein-protein and drug-drug interaction corpora, outperforming Bi-LSTM and CNN models on extracting these relations, demonstrating the complementary use of CNNs and RNNs in this relation extraction task. 

Several papers used RNNs in order to sequentially label entities of interest and determine their relations. One such work \cite{munkhdalai2018clinical} compared SVMs, RNNs, and a rule induction system, and found that while the SVM model performs the best with an F1 score of 89.1\%, a BiLSTM model had the best performance in extracting relations among RNN models for an adverse drug event (ADE) detection task. Another work \cite{sahu2018drug} used a BiLSTM model with attention to extract drug-drug interactions. Dandala, Joopudi, and Devarakonda \cite{dandala_2019} used a BiLSTM-CRF network to label entities, and another BiLSTM network with attention to assign relations. They found that in jointly modeling both tasks, their performance improved from 0.62 to 0.65 F-measure. Christopoulou et al. \cite{christopoulou_2019} employed a similar BiLSTM-CRF model for intra-sentence relation extraction, but used a Transformer-based model to capture longer dependencies in modeling inter-sentence relations. 

In another relation extraction task involving BioBERT, Alimova and Tutubalina \cite{ALIMOVA2020103382} used a random forest classifier with numerous features, such as distance-based features, word-level features, embeddings, and knowledge-base features, for relation extraction framed as a classification task, and compare it to fine-tuned BERT, BioBERT, and Clinical BERT. They improved previous results by 3.5\% on F-measure on a relation extraction task. In a protein-protein interaction (PPI) detection task, another work fine-tuned BioBERT on PubMed abstracts automatically labeled with PPI types \cite{elangovan2020assigning}. Their general approach allowed them to text mine 3,253 new typed PPIs from PubMed abstracts.

The goal of event extraction is to detect different events of interest and their properties. Usually, events have a verb indicating a specific event that connects multiple entities, which makes this task more difficult than relation extraction. A few works have studied models suitable for both relation and event extraction. B{\"\j}orne and Salakoski \cite{bjorne-salakoski-2018-biomedical} developed a CNN-based model where the input is encoded in part using dependency path embeddings. The addition of deep parsing in the systems helped model both relation and event arguments. Later, inspired by the Transformer model, ShafieiBavani et al. \cite{shafieibavani-etal-2020-global} presented a model incorporating multi-head attentions and convolutions in order to jointly model relation and event extraction. In this case, multi-headed attention helps in modeling the dependencies present in event and relation structure. Their model achieved state-of-the-art on 10 different biomedical information extraction corpora, including event and relation extraction tasks.

\subsection{Medication Information Extraction}

EHRs are a trove of vitally important information pertaining to medications. Medications and prescriptions, as some of the few actionable decision outcomes of a clinical encounter, are critical in healthcare quality analysis and in clinical research. They are often recorded in EHRs as free text, limiting access for other applications without pre-processing. Early work relied on rule-based approaches. For example, the MedEx system by Xu et al. \cite{xu2010medex} applied lookup, regular expressions, and rule-based disambiguation components. MedEx was developed with discharge summaries, and was roughly transferable to outpatient clinical visit notes in identifying medication information.

More recent work on this task has focused on expanding accessibility and transitioning toward deep learning techniques. The Clinical Language Annotation, Modeling, and Processing (CLAMP) toolkit by Soysal et al. \cite{soysal2018clamp} provided a GUI for end-users to customize their NLP pipelines for individual applications. Inspired by this system, Amazon invested heavily in its Amazon Comprehend Medical system \cite{bhatia2019comprehend}. This system performs NER to extract information about a patient's anatomy, medical condition, medications (including name, strength, and dosage) and more. It encodes the EHR texts using two LSTMs, then extracts concepts with a tag decoder. In specific, the proposed conditional softmax decoder in Amazon Comprehend Medical outperforms the best model (bidirectional LSTM-CRF) by over 1.5\% F1 score in negation detection in the 2010 i2b2 challenge \cite{stubbs2015automated}. The ease of use of these systems enables users with less technical background to make use of these NLP techniques to improve clinical outcomes.

Like Amazon Comprehend Medical, Mahajan et al. \cite{mahajan2020extracting} also used NER in their model for medication dosage extraction. Theirs, however, is the first to automatically compute daily dosage from medication instructions in the form of unstructured text. They used a BERT-based NER system to extract features related to medications, like names, frequencies, administration route, and dosage, which the model then uses to calculate the daily dosage of each medication.



%% file: 5_generation_summarization.tex
\section{Generation}
In this section, we introduce recent breakthroughs of deep learning models on generation tasks: clinical text generation, summarization, and medical language translation.

\subsection{EHR Generation}

Natural Language Generation (NLG) is one of the central components of an NLP pipeline. In the context of EHR, NLG is used to create novel clinical text from existing clinical documents \cite{huske2003text}. Generation is extremely important in the medical research domain, due to the difficultiy from EHR's accessibility and confidentiality. Making artificially created data available to researchers makes it easier to avoid privacy issues. 



RNN and LSTM based seq2seq models are commonly used in generative tasks. In an earlier paper, Lee \cite{Lee-EHR-gen} introduced an RNN-based encoder-decoder framework to generate artificial chief complaints in EHRs. In this model, the encoder takes into account many patient variables, like age, disposition, and diagnosis, and decodes them into a a chief complaint in text form. They found that the generated complaints are largely epidemiologically valid, and preserve the relationships between the diagnoses and chief complaints. 
This model has applications in clinical decision support, disease surveillance, and other data-hungry tasks. In a more domain-specific setting, Hoogi et al. \cite{mammography_gen} focused on generating artificial mammography reports via an LSTM architecture trained on real mammography reports. Such models take an image like an X-ray as input and generate a description of what it shows. Their work applied a Turing test in which they ask a radiologist to distinguish between real and generated reports; the radiologist classified real reports correctly 86\% of the time, and fake ones as real 75\% of the time, suggesting that the fake reports are of acceptably high quality. They showed that augmenting real data with generated data significantly improves performance in a downstream benign/malignant breast tumor classification task by up to 6\% on the accuracy. Similarly, Melamund and Shivade \cite{melamud-shivade-2019-towards} used an LSTM language model to generate a synthetic dataset that mimicked the style and content of the original dataset. They introduced a new notes generation task, in which the synthetic notes will be generated based on real de-identified clinical discharge summary. To benchmark this task, they collected MedText dataset that contains about 60k clinical notes extracted from MIMIC-III.

Transformers have also been used for medical text generation. Amin-Nejad et al. \cite{amin-nejad-etal-2020-exploring} introduced a Transformer-based model. Using the MIMIC-III database, they model text generation with information of the patient and the ICU stay as input, and the textual discharge summary as output. They compare the generation capacity of the vanilla Transformer model to that of GPT-2 \cite{radford2019language}. Evaluation shows that Transformer achieves BLEU score of 4.76 and ROUGE-2 score of 0.3306, while GPT-2 only achieves 0.06 and 0.1350. Moreover, they showed that the augmented dataset generated by the Transformer model is able to beat other baselines downstream tasks including readmission prediction and phenotype classification. 

In addition to seq2seq and Transformer models, Generative Adversarial Networks (GANs) \cite{ian2014generative} are also widely used for synthetic EHR generation. Choi et al. \cite{choi2018generating} proposed an approach to synthetic EHR generation via a medical Generative Adversarial Network (medGAN). This model generates high-dimensional multi-label discrete variables found within an EHR, such as medications, diagnoses, and procedures. Via a combination of an autoencoder and a GAN, the model is trained to learn the distribution of these discrete high-dimensional EHR variables. The autoencoder is trained to project real samples to a low dimentional space, and project back into the original feature space, acting as a feature extractor;  the discriminator then makes judgments on source data and generated data passed from the generator and decoder. Except for a few outliers, a trained medical professional found records from medGAN and the source data relatively indistinguishable. Baowaly et al. \cite{baowaly2019gan} improved upon the existing medGAN method, proposing two variational models: medWGAN and medBGAN. The medWGAN model uses Wasserstein GAN with gradient penalty (WGAN-GP) \cite{gulrajani2017improving} model as the generative network, which employs gradient penalties to overcome sample quality challenges; and medBGAN uses the boundary-seeking GAN (BGAN) \cite{hjelm2017boundary} model, where the generator is trained to create samples on the decision boundary of the discriminator. The synthetic EHR data generated by the three models were compared, and evaluation shows that both proposed models outperformed medGAN by up to 0.1 on the F1 score, and medBGAN performs the best.

Another typical application of generation is captioning medical images in medical reports. Writing medical reports is a tedious task for experienced radiologists, and challenging for newer radiologists who do not yet possess the requisite skills and experience. To automate the progress of writing reports, the generation model is trained to generate a paragraph description of the image, both accurately identifying all abnormalities and generating complex sentences, which are more informative and complicated than the usual natural image captions. Li et al. \cite{li2020auxiliary} proposed the Auxiliary Signal-Guided Knowledge Encoder-Decoder (ASGK) model, which attempts to mimic the work patterns of radiologists. They took advantage of two types of auxiliary signals: the internal fusion features and external medical linguistic information. The first type of feature is generated from auxiliary region features and global visual features, while the second type of feature is extracted from a large-scale medical textbook corpus. The medical graph and natural language decoders are pretrained with external auxiliary signals to memorize and phrase medical knowledge, and then trained with internal signals to support the graph encoding which integrates prior medical knowledge and visual and linguistic information. The AGSK model outperforms other state-of-the-art methods in report generation and tag classification on the CX-CHR dataset and a new COVID-19 CT report dataset, and it achieves a gain of up to 4 in BLEU score and up to 8\% on human evaluation Hit Rate.

To generate more complete and consistent radiology reports,
Miura et al. \cite{miura2020improving} proposed two reward functions to improve text quality. The first encourages systems to generate entities that are consistent with the reference; the second applies NLI to encourage these entities to be described in inferentially consistent ways.
They optimized these two rewards in a reinforcement learning framework, leading to a significant improvement over report generation baselines with up to 64.2\% gain on selected clinical metrics.

\subsection{Summarization}

Nurses, doctors and researchers deal with massive EHRs on a daily basis. Text summarization, one of the fundamental tasks in NLP, could reduce their workloads by condensing documents into brief, readable summaries \cite{mishra2014text}. Common summarization domains include news articles \cite{erkan2004lexrank,see2017get,fabbri2019multi}, scientific papers \cite{abu2011coherent,yasunaga2019scisummnet}, and dialogues \cite{zechner2001automatic}. Recent work in summarization has also looked at documents in the EHR domain. We group these pieces of work into two categories: extractive summarization and abstractive summarization. Extractive summarization selects short, discrete chunks (words, sentences, or passages) directly from the source text in order to express its most salient content as the summary. Abstractive summarization is used to generate a summary that captures the salient concepts or ideas of the source text in clear language, possibly with novel words or sentences that do not exist in the source text. 




\subsubsection{Extractive Summarization}


Extractive summarization on EHR texts has long been an interesting task. Portet et al. \cite{portet2007automatic} proposed one of the first attempts in EHR extractive summarization. Their model was designed for neonatal intensive care data, which consists of both free text and discrete information (e.g. equipment settings and drug administration). Handling these diverse types of data in one model is a challenging task, and human evaluators found model summaries to be unhelpful. For this reason, most subsequent attempts handled only textual data. Recently, Moradi et al. \cite{moradi2019small} proposed a graph-based model which ranks sentences by the important biomedical concepts they share. To circumvent the shortage of labeled training data, Liu et al. \cite{liu2018unsupervised} performed extractive summarization based on the intrinsic correlation between EHRs within a disease group to generate pseudo-labels.

There have been limited attempts for applying deep models for extractive summarization. The work from Alsentzer and Kim \cite{Alsentzer2018extractive} was the first to apply neural networks for EHR summarization. They provided an upper bound on extractive summarization on EHR discharge notes. An LSTM model is proposed to label topics in the history of present illness (HPI) notes with am F1 score of 0.876. Besides, they showed that  the model is able to create a dataset for evaluation purpose for extractive summarization methods. 
Liang et al. \cite{liang-etal-2019-novel-system} further proposed a clinical note processing pipeline and evaluated it on a disease-specific extractive summarization task on clinical notes. For the summarization model, they compared with a linear SVM, a linear chain CRF model in which each note is modeled as a sequence, and a simple CNN-based model. The CNN model outperforms the other two by up to 0.12 on the F1 score even with limited labeled data.

Unlike generic summarization as described so far, query-based summarization aims to produce a summary that is relevant to a given query. Applying query-based summarization on EHRs is extremely helpful in practice. For example, a physician may want to conduct a quick search for relevant medical information, and a summarized version of EHR can make an efficient job.  McInerney et al. \cite{mcinerney2020query} designed a query-focused extractive summarization model that selects the sentences that are the most relevant to a potential diagnosis. Because no large corpus of EHRs with extractive summaries exists, they used a distant supervision framework that extracts ICD diagnosis codes from future visits. They trained a Transformer-based neural network to select the summary sentences from an EHR and use them to predict future diagnoses, formulating the task as a sentence classification task and report precision, recall and F1. Another model that produces extractive summaries based on queries by Molla et al. \cite{molla2020query} directly compares the query sentence with each candidate sentence from the EHR. They applied different encoding methods including BERT, BioBERT and various deep model architectures (not pre-trained, pre-trained and Siamese network). However, they found that applying BioBERT did not improve the results. A strong benefit of this system is that it is capable of performing multiple-document summarization, rather than being limited to a single document.

\subsubsection{Abstractive Summarization}

\begin{figure}
\centering
\includegraphics[width=\linewidth]{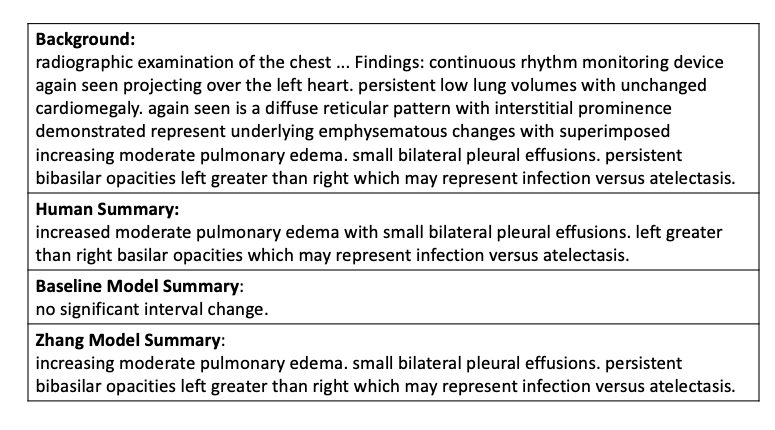}
\caption{Example result of Zhang's RL-based abstractive summarizer.\cite{zhang2020optimizing} It is capable of achieving near-human performance.}
\label{fig:abssumm_ex}
\end{figure}


Recent research of abstractive summarization has been focused on applying seq2seq and similar models to summarize radiology reports. The first work that applies seq2seq to automate the generation of radiology impressions was proposed by Zhang et al. \cite{zhang2018learning}. The model learns to encode the \lq\lq Background\rq\rq\space section in the report to guide decoding as abstractive summary. Later, \cite{macavaney2019ontologyaware} further replaced the seq2seq model by a pointer-generator model \cite{see2017get} with up to 10 improvement ROUGE-1 score and 9 improvement ROUGE-2 score compared with traditional LSA \cite{steinberger2004using} and LexRank \cite{erkan2004lexrank} models.


Reinforcement learning (RL) has also been used. In the model proposed by Zhang et al.\cite{zhang2020optimizing}. They aim to optimize an RL objective that balances the model summary's factual accuracy, linguistic likelihood, and overlap with the target summary. A factual correctness score is computed between the model summary and the CheXpert labeler \cite{irvin2019chexpert}, which extracts fact variables from a source radiology report. The results show that with a pointer-generator model \cite{see2017get} as the baseline, applying reinforcement learning leads to an improvement of roughly 3-4 on ROUGE scores.

A separate area of research is concerned with summarization of questions and their answers, there are a few benchmark datasets introduced for this topic. Abacha and Demner-Fushman \cite{abacha2019on} introduced a corpus of summarized consumer health questions and used it to train an effective pointer-generator network for abstractive summarization. Similarly, Savery et al. \cite{savery2020question} developed a dataset of common consumer health questions, their answers, and summaries of their answers. To benchmark this dataset, they evaluated state-of-the-art summarization models including a BiLSTM model, pointer-generator networks \cite{see2017get} and BART \cite{lewis2020bart}.

\subsection{Medical Language Translation}

The understanding of EHRs is limited to most readers except for professionals since they include esoteric medical terms, abbreviations, and they exhibit a unique structure and writing style. Medical language translation is the task of converting medical texts to a style that is more easily understandable by laypeople. For example, a term like \lq\lq peripheral edema\rq\rq\space may be replaced with or tagged as \lq\lq ankle swelling.\rq\rq\space 

Only a limited amount of research has focused on EHR simplification. Weng et al. \cite{weng2019unsupervised} performed unsupervised text simplification for clinical notes. Using unsupervised methods helps circumvent the shortage of texts that have been manually annotated with simplified versions. They use skip-gram embeddings learned from two clinical corpora: MIMIC-III, which includes a large amount of medical jargon, and MedlinePlus,\footnote{\url{https://medlineplus.gov/}} which is oriented towards the layperson. A Bilingual Dictionary Induction model is used to align these embeddings of technical and simpler terms and initialize a denoising autoencoder. This autoencoder inputs a physician-written sentence, generates a simpler translation with a language model, and uses back-translation to reconstruct the original sentence. 
On the supervised end of the spectrum, Luo et al. \cite{luo2020benchmark} introduced the MedLane dataset, a human-annotated Medical Language translation dataset, which aligns professional medical sentences with layperson-understandable expressions. It includes 12,801/1,015/1,016 samples for training, validation, and testing, respectively. They also proposed the PMBERT-MT model, which takes the pre-trained PubMedBERT \cite{gu2020domain}, and conducts translation training using MedLane.

%% file: 6_other_topics.tex
\section{Other Topics}
After covering some of the main NLP tasks as related to EHR, in this section, we discuss other topics that receive less attention but still important to EHRs and other relevant domains. These topics include question answering, phenotyping, medical dialogues, multilinguality, interpretability, and finally, applications in public health.  

\subsection{Question Answering}

Question answering (QA) is the task of interpreting natural language questions and retrieving appropriately paired answers \cite{kwiatkowski2019natural}. Open domain QA systems have had recent success with pre-trained language models \cite{karpukhin2020dense}, but these results have not carried over to biomedical QA because of the domain-specific challenges it faces. The primary reason for the lower performance of biomedical QA in EHRs is that models trained on open domain corpora have difficulty understanding the hyper-specific technical vocabulary often used in clinical settings. 


While the main difficulty is still the limitation of large-scale training data, pre-trained models play an important role. 
BioBERT \cite{lee2020biobert}, a pre-trained biomedical language model trained on PubMed articles, has been successfully adapted for QA tasks. 
For successful domain adaptation, pre-trained language models needs to be fine-tuned on biomedical QA datasets. However, biomedical QA datasets are often very small (e.g., just a few thousand samples), and creating new datasets is cost-prohibitive. So, Lee et al.\cite{tsatsaronis2015overview} first fine-tune BioBERT on large-scale general domain extractive QA datasets, and then fine-tune on the biomedical BioASQ dataset . Using this transfer learning framework, they are able to significantly outperform the basic BERT \cite{devlin2019bert} and other state-of-the-art models in the QA task, as well as other biomedical NLP tasks, improving and overcoming challenges in both range of vocabulary and dataset size. Vilares et al. \cite{vilares2019headqa} recognized the same problem with limited datasets and focus on the task of multi-choice QA, which requires knowledge and reasoning in complex domains. They introduced the HEAD-QA (Healthcare Dataset for Complex Reasoning) dataset, which is created from the Spanish government's annual specialized healthcare exams. They evaluated an information retrieval model on the Spanish HEAD-QA, as well as a translated English version (HEAD-QA-EN). The cross-lingual model performed better with an average accuracy of 34.6 vs. 32.9 across different sub-domains in unsupervised settings and 37.2 vs 35.2 in supervised settings. Being able to train with cross-lingual datasets or potentially automatically translate question-answer pairs could open the door to improvements in multilingual QA, solving dataset limitations by expanding possible source material.


\begin{figure}
\centering
\includegraphics[width=0.8\linewidth]{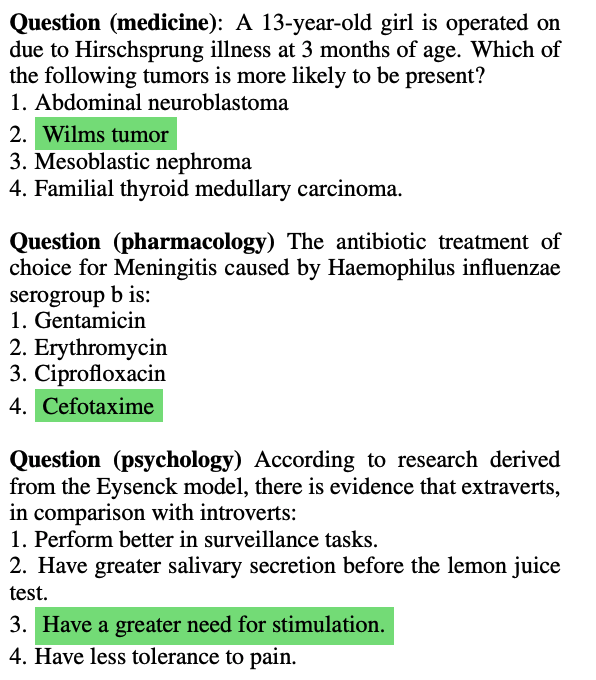}
\caption{A few samples from HEAD-QA dataset \cite{vilares2019headqa}, a multi-choice question answering healthcare dataset.}
\label{fig:headqa}
\end{figure}


It is also possible to transform the QA task to other tasks including information extraction, question similarity, information retrieval, and recognizing question entailment. 

Recently, Selvaraj et al. \cite{selvaraj2019medication} applied a QA method to extract the medication regimen (dosage and frequency for medications) discussed in a medical conversation. They formulate the Medication Regimen task as a QA task and generate questions using templates such as \lq\lq What is the <dosage/frequency> for <Medication Name>?\rq\rq\space They used an abstractive QA model based on pointer-generator networks \cite{see2017get} with a co-attention encoder to model the task, achieving ROUGE-1 scores averaging 86.40.

Question similarity and question entailment have been promising paths to solving the biomedical QA task. Many medical questions asked online often resemble \lq\lq already answered\rq \rq questions, so the goal of question entailment is to map new questions to similar answered ones. In a work by McCreery et al. \cite{mccreery2019domainrelevant}, an in-domain semi-supervised approach was proposed and tested on 3,000 medical question pairs. They pretrained BERT on the HealthTap dataset \footnote{\url{https://github.com/durakkerem/Medical-Question-Answer-Datasets}} and double fine-tuned, first on either Quora question similarity (QQP) or medical answer completion, and then on medical question pairs. The model tuned with medical answer completion reached a higher question-similarity accuracy than the QQP model, with an average difference in performance of 3.7\%.

\begin{figure}
\centering
\includegraphics[width=\linewidth]{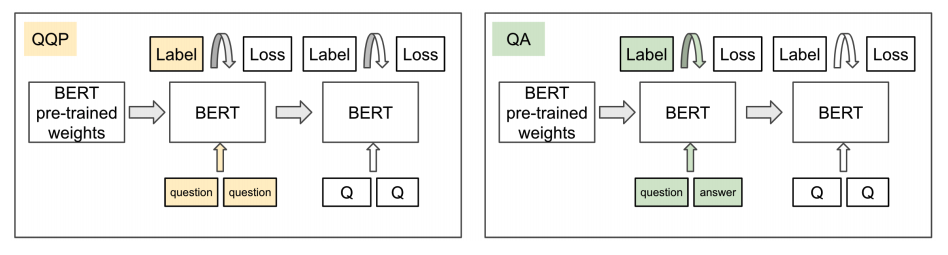}
\caption{The double finetune method using a pre-trained BERT to an intermediate task to medical question-similarity
task for two different intermediate tasks: Quora question-question pairs (left) and medical question-answer pairs
(right).\cite{mccreery2019domainrelevant} }
\label{fig:mccreery2019domainrelevan}
\end{figure}

Biomedical QA can also be solved by combining IR models with recognizing question entailment (RQE) methods. RQE is a task similar to natural language inference. RQE also extracts meaning from sentences, however it tries to create relevant relaxations of contextual and semantic constraints, such that specific questions can be related to more general and already-answered questions. A work by Abacha et al. \cite{abacha2019a} attempted two approaches to RQE: a neural network and a logistic regression classifier. The neural network performed best with the general domain NLI datasets with an average accuracy of 80.66\% on textual datasets and 83.62\% on question datasets, but logistic regression resulted in higher accuracy for the domain-specific datasets at 98.6\%. The domain-specific dataset consisted of specifically consumer-health questions which would be more applicable for general medical QA use.

Other attempts including applying paraphrasing have been proposed to improve QA systems in EHRs \cite{soni2019paraphrase,soni2020paraphrasing}. A representative work by Soni et al. \cite{soni2019paraphrase} collected 10,578 unique questions via crowdsourcing. Then a deep model consisting of a variational autoencoder and an LSTM model \cite{gupta2018deep} to create an automated clinical paraphrasing system. This model seeks to generate paraphrases without using any external resource for EHR questions. The paraphrases were evaluated on several metrics, achieving a BLEU score of 13.25, METEOR score of 21.47, and TER score of 91.93.

\subsection{Phenotyping}

Computational phenotyping is the process of extracting clinically relevant characteristics from patient data. These characteristics include physical traits, physiology, and behavior. Phenotyping is used in several areas of medical research, such as categorizing patients by diagnosis for further analysis and identifying new phenotypes \cite{Zengetal2019}. Recent techniques in computational phenotyping have replaced traditional rule-based phenotyping algorithms with scaleable NLP models. 

Zhang et al. \cite{Zhang2019UnsupervisedAO} proposed an unsupervised deep learning model to identify phenotypes in EHRs. They make use of the Human Phenotype Ontology (HPO) \cite{kohler2019expansion}, and assume the latent semantic representation of EHRs is a combination of the semantics of phenotypic abnormalities. In order to learn EHR representations, they first use an autoencoder to learn the semantic vector representations, and then merge them with representations of each phenotype. They also use a classifier to ensure the learned representations are different enough from each other. Their approach achieves competitive results, with a precision of 0.7113, recall of 0.6805, and F1 of 0.6383, and is much faster than previous phenotype identification models, taking 40.2 minutes to annotate what took other common annotation tools hours or days.

Another recent unsupervised approach, Granite \cite{henderson2017granite}, used a tensor factorization method with limited human supervision, improving on classic dimensionality reduction techniques. Granite is a robust Poisson Nonnegative tensor factorization model (NNTF) that encourages diverse and sparse latent factors. It introduces angular penalty and an L2 regularization term, reduces overlap between factors, and also introduces simplex projection on factors which results in better sparsity control. Empirical work shows Granite yields phenotypes with more distinct elements, with cosine score between 0 and 0.4, where comparables were significantly more spread out, and is better than previous tensor factoring methods at capturing rare phenotypes, as most phenotypes captured only small parts of the population (e.g. 110 v.s. 4 on the average
number of non-zero entries of diagnosis and medication
modes per phenotype) . 

Chiu et al. \cite{chiu2017ehr} introduced bulk learning to the infectious disease domain, which uses a small dataset to simultaneously train and evaluate a large amount of phenotypes. This method uses diagnostic codes as surrogate labels and trains an intermediate model based on feature abstractions. These abstractions capture common clinical concepts among multiple clinical conditions. Each disease can then be labeled by multiple clinical concepts. The training stage consists of three parts. First, base classifiers are trained to predict labels of the set of the infectious diseases. Next, predictors are aggregated through a meta-classifier, conducting a feature abstraction step which describes the extent of effect that a base model has on the prediction on a disease. In the final stage, a small subset of disease cases are collected to produce an annotation set. In effect, bulk learning serves to separate disease batches while using less data annotations. ICD9 coding evaluation shows that this method achieves a mean AUC score of 0.83, surpassing other base classifiers by 0.05-0.4. 

\subsection{Medical Dialogues}




Recent natural language understanding (NLU) research on doctor-patient dialogues has large potential implications. The two primary applications are automatic scribing and automatic health coaching.

Automatic scribing is valuable because physicians today spend hours dealing with administrative tasks like filling in information for electronic health records. The simple solution is to hire a medical scribe. Such a person takes notes about the patient-physician encounter to reduce the physician's administrative burden and ensures that documentation in the EHR is accurate and up-to-date. However, scribes are an expensive solution \cite{brady2013virtual}.
To solve this problem, NLP researchers are trying to automatically generate clinical notes from medical dialogue. One model, AutoScribe \cite{khattak2019autoscribe}, automatically parses the dialogue for entities like medications, symptoms, times, dates, referrals, and diagnoses; it then uses this information to generate a patient note. AutoScribe was evaluated on a set of 800 audio patient-clinician dialogues and transcripts, and achieved a 0.71 F1 score on this data. While AutoScribe produced strong results, the researchers noted that it could be improved by extracting more entities and training on more dialogues. 


Automatic health coaching tries to generate conversational dialogues on top of transcription and interpretation. For simple questions that do not require complex or nuanced guidance, a health coach can be a cost-effective solution.  Personal health coaches are useful but inaccessible for lower-income patients, so Gupta et al. \cite{gupta2018creating} developed an automatic health coach that sends text messages to patients. This health coach communicates important medical information, sets concrete goals, and encourages the patient to adhere to them. Another model, from Campillos-Llanos et al. \cite{llanos2020designing}, learns to simulate a patient, instead of the medical professional. They developed a \lq\lq virtual patient\rq\rq\space dialogue system with which a physician can practice clinical interactions, and found that the model produced correct replies in 74.3\% of interactions with doctors.

\subsection{Multilinguality}

There has been substantial recent interest in processing non-English medical texts, which are generally less available than ones written in English \cite{neveol2018clinical}. 

Roller et al. \cite{roller2018crosslingual} proposed a cross-lingual sequential search for candidate concepts for biomedical concept normalization. The main component of the model is a neural machine translation network trained on UMLS for Spanish, French, Dutch and German. The proposed model performs similarly to commercial translators (such as Google and Bing) on these four languages. Perez et al. \cite{perez2020crosslingual} compared the effectiveness of three approaches in automatic annotation of biomedical texts in Spanish: information retrieval and concept disambiguation, machine translation (annotating in English and translating into Spanish), and a hybrid of the two. The hybrid approach performed the best of the three, achieving an average F1 score of 0.632. 
Neural Clinical Paraphrase Generation (NCPG) is a model that casts clinical paraphrasing as a monolingual neural machine translation problem \cite{hasan-etal-2016-neural}. Using a character-level, attention-based bidirectional RNN in an encoder-decoder framework paradigm to NMT efforts, NCPG outperforms a baseline word-level RNN encoder-decoder model at the word level, with a BLEU score of 24.0 vs. 18.8, but results are more mixed at the character level with BLEU scores of 30.1 vs. 31.3. Models were evaluated on a constructed dataset which combined Paraphrase Database (PPDB) 2.0 \cite{pavlick2015domain} and UMLS to build a clinically-oriented parallel paraphrase corpus. The authors additionally show that the character-level NCPG model is superior to word-level based methods, with BLEU scores of 31.3 vs. 18.8 in the non-attention baseline and 30.1 vs. 24.0 in the attention model, likely because it tackles the out-of-vocabulary problem directly.

Multilingual BERT models are also being investigated for EHR tasks. Vunikili et al. \cite{vunikili2020clinical} studied BERT-based embeddings trained on general domain Spanish text for tumor morphology extraction in Spanish clinical reports. The model achieves an F1 score of 71.3\% on the NER task without any feature engineering or rule-based methods. Silvestri et al. \cite{Silvestri2020multilingual} investigated the Cross-lingual Language Model (XLM) \cite{NEURIPS2019_c04c19c2} by fine-tuning on English language training data and testing performance on ICD-10 code classification in an Italian language dataset of short medical notes. The XLM model shows an improvement over mBERT\footnote{The multilingual version of BERT: \url{https://github.com/google-research/bert/blob/master/multilingual.md}} in this task, achieving an accuracy of 0.983 compared to 0.968.

\subsection{Interpretability}


Many machine learning models such as random forest and bootstrapping operate like black boxes; they are not directly interpretable. However, efforts have been made to improve interpretability to make these models more helpful in practice.


One major challenge comes from the trade-off between interpretability and performance. 
Caruana et al. \cite{caruana2015intelligible} applied high-performance generative additive models with pairwise interactions to pneumonia risk prediction, and were able to reveal clinically-relevant patterns while preserving SOTA performance. The model also provides a simple mechanism to correct discovered patterns and confounding factors, as with asthma, chronic lung disease, and chest pain history in a pneumonia case study. A work by Choi et al. \cite{choi2016retain} proposed the Reverse Time Attention Model (RETAIN), which improves interpretability by using a two-layer attention model, giving the most recent visits higher attention. Evaluation on Heart failure prediction performance shows that it achieved 0.87 on the AUC score which is comparable to SOTA deep learning methods like RNNs. 
Another approach to increase interpretability was suggested by Che et al. \cite{che2015distilling}, who introduced a knowledge-distillation approach called interpretable mimic learning. The model uses gradient boosting trees to learn interpretable features from existing deep learning models including LSTM and Stacked Denoising Autoencoder. Evaluated on an ICU dataset for acute lung injury, it achieves similar or better performance than deep learning models, with AUC averaging 0.760 vs. 0.756.


\subsection{Applications in Public Health}

The COVID-19 pandemic, as a public health crisis, is impacting people's lives in multiple ways. It has also caused an information crisis, with the development of the Internet and other techniques. It is estimated that more than 23k published papers have been indexed on Web of Science and Scopus just between January 1 and June 30, 2020 \cite{teixeira2021publishing}. 
In the NLP domain, researchers are focusing on processing pandemic-generated data and working on various tasks including text classification, information retrieval, named entity recognition and knowledge discovery \cite{chen2020artificial, wang2020comprehensive, li2020mental}. While there are many relevant papers, we list here some typical works that perform NLP techniques on COVID-19 data. We limit our focus to resource works.

TREC-COVID \cite{voorhees2020trec,roberts2020treccovid} is an information retrieval shared task to promote and support research related to the pandemic. Among the participants, MacAvaney et al. \cite{macavaney2020sledge} introduced a zero-shot SciBERT-based ranking algorithm for COVID-related scientific literature. Bendersky et al. \cite{bendersky2020rrf102} presented a weighted hierarchical rank fusion approach. The approach combines results from lexical and semantic retrieval systems, pretrained and fine-tuned BERT rankers, and relevance feedback runs \cite{zhang2020covidex}. They were able to achieve SOTA performance in rounds 4 and 5 of the challenge, with a 9.2\% mean average precision gain over the next-best team. 

The Open Research Dataset Challenge (CORD-19) corpus \cite{wang20202020covidopen} is a resource collection of scientific papers on COVID-19 and coronavirus research: PubMed PMC, bioRxiv and medRxiv corpus collected from search results using \lq COVID-19 and coronavirus research\rq\space as query, in addition to the WHO corpus of COVID-19 research papers. The initial release contains 28K papers, and it has been updated frequently. This dataset promotes a number of research directions including text classification \cite{liang2020identifying}, information extraction \cite{wang2020comprehensivenamed}, knowledge graphs \cite{ahamed2020informationmining} and more. 

Finally, COVID-KG \cite{wang2020coviddrug} is a knowledge discovery framework focused on extracting multimedia knowledge elements from 25,534 peer-reviewed papers. In COVID-KG, nodes are entities/concepts and edges are relations and events among these entities. The edges are extracted from both images and texts. Specifically, the knowledge graph contains multiple types of entities and links. After construction, they are able to perform a number of tasks related to COVID and drugs, including question answering and report generation \footnote{\url{http://blender.cs.illinois.edu/covid19/visualization.html}}. 

%% file: 8_conclusion.tex
\section{Conclusion and Future Direction}

In this survey, we reviewed recent studies that show how EHR and health informatics tasks can benefit from deep NLP models. Though some of these papers partially deal with structured data, our focus was on unstructured text data for downstream EHR tasks. More specifically, we summarized recent work on the following EHR-NLP tasks: classification and prediction, representation learning, extraction, generation, as well as other topics such as question answering, phenotyping, knowledge graphs, multilinguality, medical dialogues and applications on public health. We also \hyperref[appendix]{list} some relevant datasets and existing tools to promote EHR-NLP research.

Though deep learning methods in the general NLP domain have achieved remarkable success, applying them to the biomedical field is still challenging due to the limited availability and difficulty of domain-specific textual data and the lack of interpretability of deep learning methods. One future direction in this field is better mining knowledge and information from unstructured data \cite{Esteva2019}, and a useful combination of both structured and unstructured data for better decision making and potential interpretability. Another direction could be employing transfer learning or unsupervised learning for EHR tasks to compensate for the dearth of annotated textual data. We hope that our survey will inspire readers and promote future developments in NLP for electronic health records.

%% file: appendix.tex
\section{Appendix: Datasets and Tools}\label{appendix}

\subsection{Datasets}

\subsubsection{General EHR-NLP Datasets}

\textbf{MIMIC-III}\footnote{\url{https://mimic.physionet.org/}} 
A free, publicly accessible database of de-identified medical information on patient stays in the critical care units of the Beth Israel Deaconess Medical Center between 2001 and 2012. The table NOTEVENTS contains clinical notes from over 40,000 patients. Other tables have data on mortality, imaging reports, demographics, vital signs, lab tests, drugs, and procedures. Before MIMIC-III, there were two other iterations of MIMIC used by biomedical NLP researchers.

\textbf{MIMIC-CXR}\footnote{\url{https://physionet.org/content/mimic-cxr/2.0.0/}} 
Like MIMIC-III, MIMIC-CXR contains de-identified clinical information from the Beth Israel Deaconess Medical Center. It has over 377,000 radiology images of chest X-rays. The creators of the dataset also used the ChexPert\cite{irvin2019chexpert} tool to classify each image's corresponding free-text note into 14 different labels.

\textbf{NUBes-PHI} \cite{lima2020nubes} A Spanish medical report corpus, containing about 7,000 real reports with annotated negation and uncertainty information.


\textbf{Abbrev dataset}\footnote{\url{https://nlp.cs.vcu.edu/data.html}} This dataset is a re-creation of an old dataset which contains the acronyms and long-forms from Medline abstracts. It is automatically re-created by identifying the acronyms long forms in the Medline abstract and replacing it with it's acronym. There are three subsets containing 100, 200 and 300 instances respectively. \cite{stevenson2009disambiguation}

\textbf{MEDLINE}\footnote{\url{ https://www.nlm.nih.gov/bsd/medline.html}}
A database of 26 million journal articles on biomedicine and health from 1950 to the present. It is compiled by the United States National Library of Medicine (NLM). MedlinePlus\footnote{\url{ https://www.nlm.nih.gov/bsd/medline.html}}, a related service, describes medical terms in simple language.

\textbf{PubMed}\footnote{\url{ https://www.ncbi.nlm.nih.gov/guide/howto/obtain-full-text/}} 
A corpus containing more than 30 million citations for biomedical and scientific literature. In addition to MEDLINE, these texts come from sources like online books, papers on other scientific topics, and biomedical articles that have not been processed by MEDLINE. 

\subsubsection{Task-Specific Datasets}

\textbf{BioASQ}\footnote{\url{ http://bioasq.org/}} 
An organization that designs challenges for biomedical NLP tasks. While BioASQ challenges focus primarily on question answering (QA) and semantic indexing, some use other tasks including multi-document summarization, information retrieval, and hierarchical text classification.

\textbf{BIOSSES} \footnote{\url{http://tabilab.cmpe.boun.edu.tr/BIOSSES/}} \cite{souganciouglu2017biosses} A benchmark dataset for biomedical sentence similarity estimation.

\textbf{BLUE} \footnote{\url{https://github.com/ncbi-nlp/BLUE_Benchmark}} 
Biomedical Language Understanding Evaluation (BLUE) is a collection of ten datasets for five biomedical NLP tasks. These tasks cover sentence similarity, named entity recognition, relation extraction, document classification, and natural language inference. 
BLUE serves as a useful benchmarking tool, as it centralizes the datasets that medical NLP systems evaluate on.

\textbf{Clinical Abbreviation Sense Inventory}\footnote{\url{https://conservancy.umn.edu/handle/11299/137703}}
A dataset for medical term disambiguation. In the latest version, 440 of the most frequently used abbreviations and acronyms were selected from 352,267 dictated clinical notes.

\textbf{CLINIQPARA}\cite{soni2019paraphrase} A dataset with paraphrases for clinical questions. Contains 10,578 unique questions across 946 semantically distinct paraphrase clusters. Initially collected for improving question answering for EHRs.

\textbf{i2b2}\footnote{\url{ http://www.i2b2.org/}} 
Informatics for Integrating Biology and the Bedside, or i2b2, is a non-profit that organizes datasets and competitions for clinical NLP. It has numerous datasets for specific tasks like deidentification, relation extraction, clinical trial cohort selection. These datasets and challenges are now run by Harvard's National NLP Clinical Challenges, or n2c2; however, most papers refer to them with the name i2b2.

\textbf{MedICaT}\footnote{\url{ https://github.com/allenai/medicat}}
A collection of more than 217,000 medical images, corresponding captions, and inline references, made for figure retrieval and figure-to-text alignment tasks. Unlike previous medical imaging datasets, subfigures and subcaptions are explicitly aligned, introducing the specific task of subcaption-subfigure alignment.

\textbf{MedNLI}\footnote{\url{ https://jgc128.github.io/mednli/}} 
Designed for Natural Language Inference (NLI) in the clinical domain. The objective of NLI is to predict whether a hypothesis can be deemed true, false, or undetermined from a given premise. MedNLI contains 14,049 unique sentence pairs, annotated by 4 clinicians over the course of six weeks. To download it, one first needs to get access to MIMIC-III.

\textbf{MedQuAD}\footnote{\url{https://github.com/abachaa/MedQuAD}}
Medical Question Answering Dataset, a collection of 47,457 medical question-answer pairs created from 12 NIH websites (e.g. cancer.gov, niddk.nih.gov, GARD, MedlinePlus Health Topics). There are 37 question types associated with diseases, drugs, and other medical entities such as tests. \cite{019arXiv190108079A}

\textbf{VQA-RAD}\footnote{\url{https://osf.io/89kps/}} 
A dataset of manually constructed question-answer pairs corresponding to radiology images. It was designed for future Visual Question Answering systems, which will automatically answer salient questions on X-rays. These models will hopefully be very useful clinical decision support tools for radiologists.

\textbf{WikiMed and PubMedDS} \cite{vashishth2020medtype} Two large-scale datasets for entity linking. WikiMed contains over includes 650,000 mentions normalized to concepts in UMLS. PubMedDS is an annotated corpus with more than 5 million normalized mentions spanning across 3.5 million documents.

\textbf{PathVQA}
\cite{he2020pathvqa} The first dataset for pathology visual question answering. It contains manually-checked 32,799 questions from 4,998 pathology images.

\textbf{MedQA}
\cite{jin2020disease} The first multiple-choice OpenQA dataset
for solving medical problems. The dataset is collected from professional medical board exams on three languages: English, simplified Chinese, and traditional Chinese. For the languages, there are 12,723, 34,251, and 14,123 questions respectively. 

\balance
\subsection{Tools and libraries}

\subsubsection{NLP and Machine Learning}

\textbf{Pytorch} \footnote{\url{ https://pytorch.org/}}
Pytorch is an open source deep learning library developed by Facebook, primarily for use in Python. It is a leading platform in both industry and academia.

\textbf{Scikit-learn}\footnote{\url{https://scikit-learn.org/}}
An open python library providing efficient data mining and data analysis tools. These includes methods for classification, regression, clustering, etc. 

\textbf{TensorFlow}\footnote{\url{https://www.tensorflow.org/}} Google's open-source framework for efficient computation, used primarily for machine learning. 
Tensorflow provides stable APIs for Python and C; it has also been adapted for use in a variety of other programming languages.

\textbf{AllenNLP}\footnote{\url{https://allennlp.org/}} An open-source NLP research library built on PyTorch. AllenNLP has a number of state-of-the-art models readily available, making it very easy for anyone to use deep learning on NLP tasks.
	
\textbf{Fairseq}\footnote{\url{https://github.com/pytorch/fairseq}} A Python toolkit for sequence modeling. It enables users to train models for text generation tasks like machine translation and language modeling.
	

\textbf{Gensim}\footnote{\url{https://radimrehurek.com/gensim/index.html}} A scalable, robust, efficient, and hassle-free python library for unsupervised semantic modelling from plain text. It has a wide range of tools for topic modeling, document indexing, and similarity retrieval.

\textbf{Natural Language Toolkit (NLTK)}\footnote{\url{https://www.nltk.org/}} A leading platform for building Python programs concerned with human language data. It contains helpful functions for tasks such as tokenization, cleaning, and topic modeling. 

\textbf{PyText} \footnote{\url{https://pytext-pytext.readthedocs-hosted.com/en/latest/}}  PyText is a deep-learning based NLP modeling framework built on PyTorch, providing pre-trained models for NLP tasks such as sequence tagging, classification, and contextual intent-slot models.

\textbf{SpaCy}\footnote{\url{https://spacy.io/}} A remarkably fast Python library for modeling and processing text in 34 different languages. It includes pretrained models to predict named entities, part-of-speech tags, and syntactic dependencies, as well as starter models designed for transfer learning. It also has tools for tokenization, text cleaning, and statistical modeling.

\textbf{Stanford CoreNLP}\footnote{\url{https://stanfordnlp.github.io/CoreNLP/}} A set of tools developed by Stanford NLP Group for statistical, neural, and rule-based problems in computational linguistics. Its software provides a simple, useful interface for NLP tasks like NER and part-of-speech (POS) tagging. 

\subsubsection{EHR-NLP}

\textbf{Criteria2Query} \footnote{\url{http://www.ohdsi.org/web/criteria2query/}} A system for automatically transforming clinical research eligibility criteria to Observational Medical Outcomes Partnership (OMOP) Common Data Model-based executable cohort queries. \cite{yuan2019criteria2query} The system is an information extraction pipeline that combines machine learning and rule-based methods. 

\textbf{CuiTools} \footnote{\url{http://cuitools.sourceforge.net/}}
 A package of PERL programs for word sense disambiguation (WSD)\cite{mcinnes2007using}. Its models perform supervised or unsupervised WSD using both general English knowledge and specific medical concepts extracted from UMLS.
 
\textbf{Metamap}\footnote{\url{https://metamap.nlm.nih.gov/}} A tool to identify medical concepts from the text and map them to standard terminologies in the UMLS. MetaMap uses a knowledge-intensive approach based on symbolic methods, NLP, and computational-linguistic techniques.

\textbf{MIMIC-Extract} \footnote{\url{https://github.com/MLforHealth/MIMIC_Extract}} An open source pipeline to preprocess and present data from MIMIC-III. \cite{wang2020mimic} MIMIC-Extract has useful features for analysis - for example, it transforms discrete temporal data into a time-series and extracts clinically relevant targets like mortality from the text.

\textbf{ScispaCy}\footnote{\url{https://allenai.github.io/scispacy/}} Many NLP models perform poorly under domain shift, so ScispaCy adapts SpaCy's models to process scientific, biomedical, or clinical text. It was developed by AllenNLP in 2019 and includes much of the same functionality as SpaCy.

\textbf{Unified Medical Language System (UMLS)}\footnote{\url{https://www.nlm.nih.gov/research/umls/}} UMLS is a set of files and software that provides unifying relationships across a number of different medical vocabularies and standards. Its aim is to improve effectiveness and interoperability between biomedical information systems like EHRs. It can be used to link medical terms, drug names, or billing codes across different computer systems.